# An Architectural Approach to
# Ensuring Consistency in Hierarchical Execution

**Robert E. Wray**                                         WRAYRE@ACM.ORG
*Soar Technology, Inc., 3600 Green Court, Suite 600*
*Ann Arbor, MI 48105 USA*

**John E. Laird**                                          LAIRD@UMICH.EDU
*The University of Michigan, 1101 Beal Avenue*
*Ann Arbor, MI 48109 USA*

## Abstract

Hierarchical task decomposition is a method used in many agent systems to organize agent knowledge. This work shows how the combination of a hierarchy and persistent assertions of knowledge can lead to difficulty in maintaining logical consistency in asserted knowledge. We explore the problematic consequences of persistent assumptions in the reasoning process and introduce novel potential solutions. Having implemented one of the possible solutions, *Dynamic Hierarchical Justification*, its effectiveness is demonstrated with an empirical analysis.

## 1. Introduction

The process of executing a task by dividing it into a series of hierarchically organized subtasks is called *hierarchical task decomposition*. Hierarchical task decomposition has been used in a large number of agent systems, including the Adaptive Intelligent Systems architecture (Hayes-Roth, 1990), ATLANTIS (Gat, 1991a), Cypress (Wilkins et al., 1995), the Entropy Reduction Engine (Bresina, Drummond, & Kedar, 1993), the Procedural Reasoning System (Georgeff & Lansky, 1987), RAPS (Firby, 1987), Soar (Laird, Newell, & Rosenbloom, 1987; Laird & Rosenbloom, 1990), and Theo (Mitchell, 1990; Mitchell et al., 1991), and is a cornerstone in belief-desire-intention-based agent implementations (Rao & Georgeff, 1991; Wooldridge, 2000). Hierarchical task decomposition helps both an agent's knowledge developer and the agent itself manage environmental complexity. For example, an agent may consider high-level tasks such as "find a power source" or "fly to Miami" independent of low-level subtasks such as "go east 10 meters" or "turn to heading 135." The low-level tasks can be chosen dynamically based on the currently active high level tasks and the current situation; thus the high-level task is progressively decomposed into smaller subtasks. This division of labor simplifies the design of agents, thus reducing their cost. Additional advantages of hierarchical task decomposition include knowledge sharing (a low-level subtask can be invoked for many different high-level procedures), modularity (the decomposition helps insulate subtasks from interaction with other knowledge) and the naturalness of this representation (Simon, 1969).

Without careful design, it can be difficult to ensure consistent reasoning in agents employing hierarchical task decompositions. By "consistency," we mean that reasoning does not lead to a set of assertions that contains a contradiction. Ensuring consistency becomes much more difficult to solve – and thus more costly – as the complexity of an agent's knowledge grows. Although this problem can be solved through careful design of agent knowledge, such an approach requires an understanding of all possible interactions in the hierarchy. Thus, the correctness of this solution depends on the skill and vigilance of the knowledge engineer. Our bias is to seek solutions in which the operation of an agent's primitive memories and processes are structured to ensure inconsistencies do not arise. Thus, we will prefer architectural solutions to knowledge-based ones. Architectural solutions can guarantee consistency for all tasks and domains, reducing brittleness due to omissions in task knowledge. Further, while developing an architectural solution may be costly, it should be less costly than repeatedly developing knowledge-based solutions for different domains.

The following sections describe the inconsistency problem and introduce a space of solutions to the problem, including two novel solutions. Through both theoretical and empirical analysis, one of the new solutions, *Dynamic Hierarchical Justification*, is shown to provide an efficient architectural solution to the problem of ensuring reasoning consistency in hierarchical execution.

## 2. Maintaining Reasoning Consistency in Hierarchical Agents

This section describes the inconsistency problem in greater detail. We review methods for ensuring consistency in non-hierarchical systems and discuss the limitations of these approaches in hierarchical systems.

### 2.1 Consistency in Non-hierarchical Systems

Truth maintenance systems (TMSs) are often used to maintain consistency in non-hierarchical systems (Doyle, 1979; McDermott, 1991; Forbus & deKleer, 1993). An inference engine uses domain knowledge to create two different kinds of assertions of knowledge in an agent's knowledge base: *assumptions* and *entailments*. The inference engine *enables* assumptions that it has decided to treat as being true, without requiring that the assertion be justified. Agents often treat environmental percepts as assumptions or "unquestioned beliefs" (Shoham, 1993). Entailments are justified assertions. A data structure, the *justification*, captures the reasons for asserting the entailment. When the reasons no longer hold (the entailment is no longer justified), the TMS retracts it from the set of asserted beliefs. Thus, a TMS automatically manages the assertion and retraction of entailments as an agent's situation changes, ensuring all entailments are consistent with the external environment and the enabled assumptions.

Careful construction of the domain knowledge is required to ensure that no enabled assumptions are contradictory. For example, if some assumption is inconsistent with the current input, then the agent must have domain knowledge that recognizes the situation and removes the assumption. Thus, when an agent utilizes a TMS, the problem of maintaining consistency in reasoning is largely one of managing assumptions through the agent's domain knowledge.

Assumptions often reflect hypothetical reasoning about the world (hence "assumptions"). However, assumptions can be used to represent any persistent feature. Although researchers have explored structuring the external environment to provide persistent memory (Agre & Horswill, 1997), internal, persistent memory is usually necessary in agent domains. For example, persistence is required for hypothetical reasoning, nonmonotonic revisions of assertions (such as when counting), and remembering.

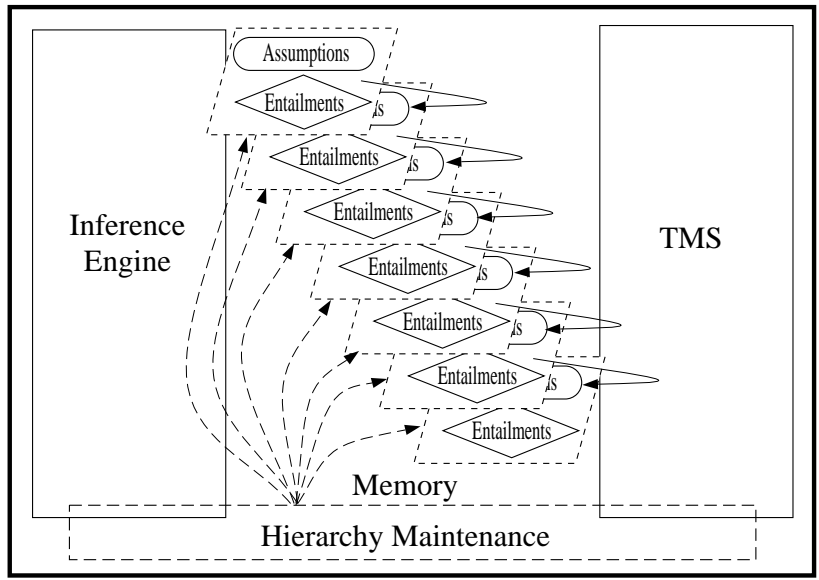

Figure 1: A hierarchical agent.

## 2.2 Truth Maintenance in Hierarchical Agents

The TMS agent framework introduced above can be extended to hierarchical agent architectures. In such an agent, the inference engine and TMS are more or less identical to those of a non-hierarchical agent. When the agent initiates a new subtask via dynamic hierarchical task decomposition, it also creates a new database that will contain assumptions and entailments specific to the subtask. Further decomposition can result in a stack of subtasks, each containing entailments and assumptions specific to the subtask, as shown in Figure 1. We consider the creation and deletion of these distinct databases of assertions the *sine qua non* of a hierarchical architecture. The architecture decomposes the task not only by identifying relevant subtasks, but also by dynamically organizing its memory according to the current decomposition.

A new system component, "hierarchy maintenance," is responsible for creating and destroying the subtask databases when subtasks begin and terminate. When a subtask is achieved (or determined to be no longer worth pursuing), hierarchy maintenance responds by immediately removing all the assertions associated with the subtask. This function is of central importance in hierarchical architectures because it allows the agent to automatically retract all assertions associated with a terminated subtask, requiring no agent knowledge to "clean up" or remove individual assertions associated with the terminated subtask. The hierarchy maintenance component can efficiently remove the assertions because they are (conceptually) located in a distinct unit, the subtask database.

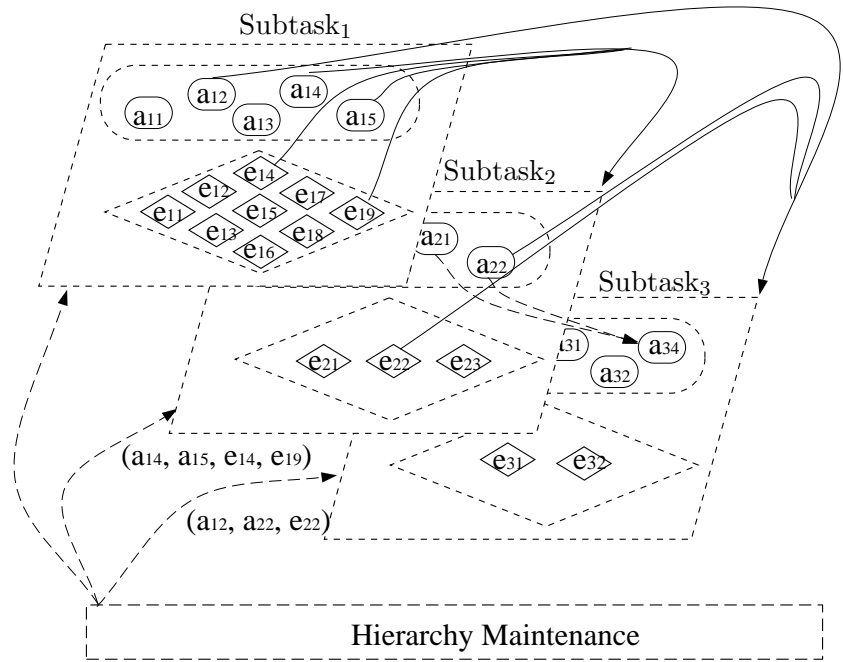

Figure 2: An example of hierarchy maintenance. Assumptions ("a") and entailments ("e") are asserted within subtasks.

An agent's hierarchy maintenance function can be employed to help maintain consistency, illustrated notionally in Figure 2. The agent architecture identifies assertions at each of the higher levels in the hierarchy that led to a new subtask. These assertions together form a subtask "support set." In Figure 2, assertions $a_{14}$, $a_{15}$, $e_{14}$, and $e_{19}$ form the support set for $Subtask_2$ while $a_{12}$, $a_{22}$, $e_{22}$ "support" $Subtask_3$. These support sets, in effect, form justifications for subtasks in the hierarchy. When an assertion in a support set is removed (e.g., $a_{22}$), the agent responds by removing the subtask ($Subtask_3$). While not all hierarchical architectures use architectural processes to create and destroy subtask databases, this example illustrates how an architectural hierarchical maintenance function can be realized via a process similar to that of justification in truth maintenance.

Within a specific subtask, reason maintenance can go on as before. However, the hierarchical structure adds a complication to the maintenance of logical consistency. Assumptions at some level in the hierarchy can be dependent on entailments and assumptions in higher levels of the hierarchy.[1] This dependence relationship is suggested in Figure 1 by the curved lines extending from one subtask to the one below it. Higher levels of the hierarchy form a "context" for reasoning in the local subtask.

For execution agents embedded in dynamic domains, the hierarchical context may change at almost any time. The changing context is not problematic for entailments; the TMS can readily determine dependent context changes and retract affected entailments. However, changes in higher levels of the hierarchy (such as those deriving from inputs) may also invalidate the assumptions of lower levels. Without any additional architectural mechanisms, domain knowledge is required to ensure consistency among assumptions and the hierarchical context as in non-hierarchical systems. The domain knowledge for ensuring consistency in the assumptions is complicated by the necessity of spanning multiple (possibly many) subtasks. We refer to such knowledge as "across-level" consistency knowledge. As described in further detail below, identifying and creating across-level consistency knowledge is a tedious, costly, and often incomplete process. Across-level knowledge must explicitly consider the interactions between different subtasks (in different levels of the hierarchy), rather than focus solely on the local subtask, compromising the benefit of the hierarchical decomposition.

---

1. Assumptions in a lower level subtask are always at least *indirectly* dependent on the higher level assertions. This observation will be exploited in Section 3.3.

Before continuing, we note that hierarchical architectures should be contrasted with hierarchical task network (HTN) planners (Sacerdoti, 1975; Erol, Hendler, & Nau, 1994) and execution-oriented systems that use HTN representations, such as DECAF (Graham & Decker, 2000) and RETSINA (Sycara, Decker, Pannu, Williamson, & Zeng, 1996). A planning problem for an HTN planner is represented by an initial task network that can consist of primitive and non-primitive tasks. The planner uses operators to find a plan to solve the tasks. Methods allow the planner to match non-primitive tasks with other task networks that describe how to accomplish the task; thus, methods enable hierarchical decomposition of the planning problem into a family of connected task networks.

The main difference between HTN systems and hierarchical architectures is that the planner represents its plan in a single global state. That is, while methods represent decomposition steps, the hierarchical structure of an evolving plan is represented in a blackboard-like database that does not also reflect the structure of the decomposition. The following sections discuss problems and especially solutions that depend on the hierarchical *organization* of asserted knowledge during execution, in addition to a hierarchical task decomposition encoded as an agent's task knowledge. Thus, the following will not be generally applicable to HTN-based execution systems. However, HTN systems need to address the inconsistency problem; Section 5.1.1 examines the consequences of global state with respect to inconsistency arising from persistence in a hierarchy.

## 2.3 Failing to Respond to Relevant Changes in Hierarchical Context

As mentioned in the introduction, when an agent fails to respond to a relevant change in its hierarchical context and leaves a now-inconsistent assumption enabled, the resulting behavior can become irrational; that is, not consistent with its knowledge. This section explores how such irrational behavior can arise with several illustrative examples.

### 2.3.1 The Blocks World

We use a variant of the blocks world to illustrate the inconsistency problem in a domain familiar to most readers. This domain is an execution domain rather than a planning domain, which we call the "Dynamic Blocks World" to reflect this difference from the static blocks world used in planning. We assume the agent has knowledge to build an ordered tower (**1-on-2-on-3**) without resorting to planning and uses hierarchical task decomposition to determine what actions to take as it builds the tower.

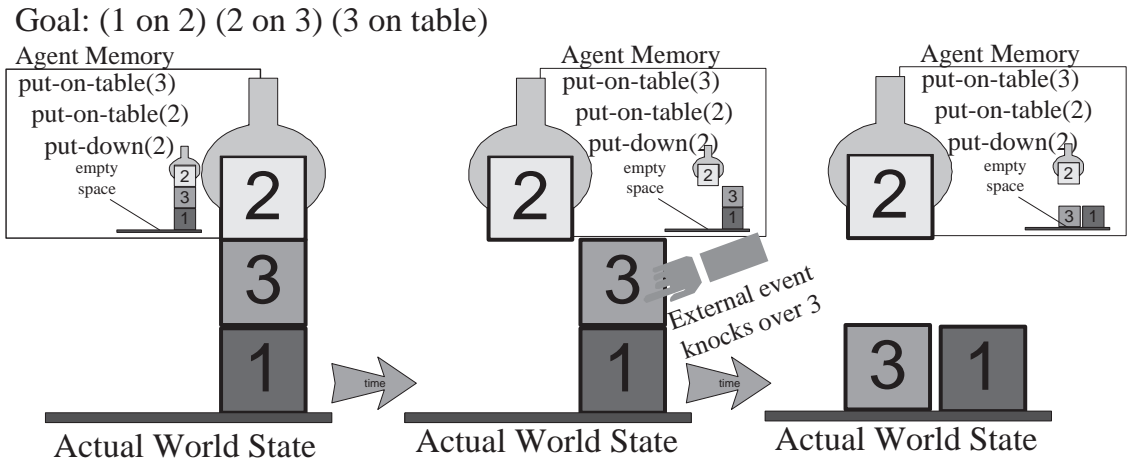

Figure 3: Failing to respond to relevant changes in hierarchical context in the Dynamic Blocks World.

In Figure 3, the agent is placing **block-2** on the table, in order to reach **block-3** and begin the goal tower. The `put-down` subtask finds an empty location on the table. The agent places the `empty` assertion in the memory associated with the `put-down` subtask. In the figure, the space immediately to the left of the gripper was chosen. Whether or not a space is empty may not be directly observable but may need to be inferred from a number of other facts in the domain and stored as an assumption in memory. Assume the `empty` assertion is an assumption. Now, assume **block-3** is suddenly placed underneath **block-2**. The result is an inconsistency between the assumption (the location is a good place to put **block-2**) and the hierarchical context (the location is no longer a good place to put the block on the table).

If the agent fails to recognize that **block-3** has moved, it will attempt to put **block-2** into the same location occupied by **block-3**. This behavior is irrational, or not consistent with the agent's goals and knowledge (assuming the agent has knowledge that indicates that blocks should not be placed in positions already occupied by other blocks). The inconsistency arises because the agent has failed to recognize its previously-derived assumption (`empty`) is no longer true in the current situation.

Although this example may appear contrived, this specific situation arose in an experimental system developed to explore architecture and learning issues. Of course, it is possible in such a simple domain to reformulate the task such that the problem does not occur. This reformulation of the task via changes to or additions of knowledge is *exactly* the solution we wish to avoid. That is, we desire that the architecture guarantee consistency between the hierarchical context and local assumptions such that the architecture provides *a priori* constraints (guidance) in the knowledge development process and increased robustness in execution (via consistency). The conclusion returns to this example to describe how an architectural solution to the inconsistency problem solves this particular problem – without requiring any reformulation of the agent's task knowledge.

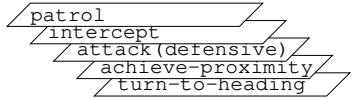

Figure 4: Decomposition of behavior into subtasks.

### 2.3.2 TACAIR-SOAR

TacAir-Soar agents pilot virtual military aircraft in a complex, real-time computer simulation of tactical combat (Tambe et al., 1995; Jones et al., 1999). The TacAir-Soar domain is only indirectly accessible (each agent uses simulated aircraft sensor models and can perceive only what a pilot in a real aircraft would sense), nondeterministic (from the point of view of the agent, the behavior of other agents cannot be strictly predicted or anticipated), non-episodic (the decisions an agent makes early in the simulation can impact later options and capabilities), dynamic (the world changes in real time while the agent is reasoning), and continuous (individual inputs have continuous values). Domains with these characteristics are the most difficult ones in which to create and apply agents (Russell & Norvig, 1995).

The domain knowledge of TacAir-Soar agents is organized into over 450 subtasks; during execution, the resulting hierarchical task decomposition sometimes reaches depths of greater than 10 subtasks. Each agent can have one of several different mission roles, among them flying a patrol mission, and acting as a partner or "wing" to some other agent's "lead."

Consider a pair of planes on patrol, which have been given specific instructions for engaging enemy aircraft. When enemy aircraft enter the patrol area, the lead agent decides to `intercept` the aircraft. The lead then decomposes the intercept into a series of situation-dependent subtasks, which themselves may be further decomposed. For example, Figure 4 shows that the complex task of intercepting an enemy aircraft has been decomposed into a decision to turn the agent's aircraft to a specific heading. The agent turns to this heading in order to get close enough to the enemy agent (via `achieve-proximity`) to launch an attack.

Assume three different kinds of attack can be chosen for an `intercept`. The first tactic (`scare`) is to engage and attempt to scare away enemy planes without using deadly force. This tactic is selected when the rules of engagement specify that deadly force should not be used, regardless of the number of aircraft in the area. One of the remaining two tactics will be chosen when deadly force is allowed. `Offensive attack` is appropriate when friendly planes outnumber or equal enemy planes. `Defensive attack` is used when enemy planes outnumber friendly planes.

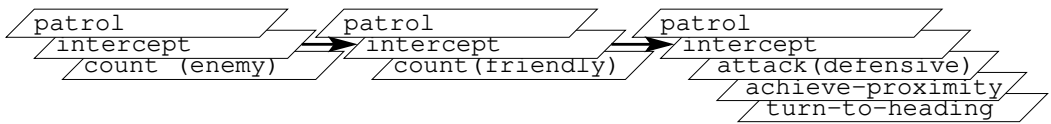

Figure 5: Trace of behavior leading to intercept tactic in TacAir-Soar.

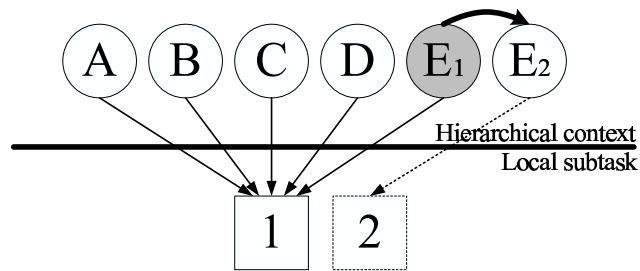

Figure 6: Inconsistency due to persistence.

Choosing between offensive and defensive attack requires counting the current aircraft in the area. Figure 5 shows the evolution of an executing example decomposition. The agent must count *relevant* enemy and friendly planes. Determining a plane's "side" and its relevance to the count often requires remembering and is sufficiently complex that entailment of the count is not possible. For instance, non-combatant aircraft should not be counted, requiring some reasoning about the type of each aircraft. If the agent determines that enemy planes outnumber friendly ones, the agent selects `defensive-attack`, leading to further decomposition.

What happens if an enemy plane flees, thus reducing the actual count of relevant enemy planes by one? The count maintained by the agent is now invalid. Standard TMS mechanisms are insufficient because the count was asserted as an assumption. If the actual number of enemy and friendly planes is now equal, then the agent should switch its tactic to offensive attack. Continuing the defensive attack is not consistent with the agent's knowledge. Additionally, other "friendly" agents participating in the attack may base their behavior on the expectation that the agent is pursuing an offensive attack. Thus the agent needs to recognize the inconsistency and remove the current count.

Figure 6 presents a conceptual illustration of the problem. Assumptions are represented as squares, entailments as circles. The horizontal line represents a hierarchical relationship between the assertions (i.e., assumptions and entailments) in the hierarchical context (above the line) and assertions in a local subtask (below the line). The arrowed lines represent dependence in the creation of an assertion. As in the previous examples, some reasoning in the subtask may require persistence, leading to the creation of an assumption such as assumption **1**. However, the persistent assertion may still depend on other assertions. This work focuses on the dependent assertions in the higher level context, such as **A**, **B**, **C**, **D**, and $\mathbf{E_1}$ in the figure.

Suppose the world changes so that $\mathbf{E_1}$ is retracted from memory and $\mathbf{E_2}$ is asserted. Assumption **1** remains in memory. If $\mathbf{E_2}$ would not also lead to **1** (e.g., it could lead to some new assumption **2**, as shown), then **1** is no longer justified and may not be consistent with the higher level context. Whether or not this potential inconsistency among the assertions leads to inconsistent behavior depends on the use of assumption **1** in later reasoning.

## 3. Solutions

Our goal is to develop architectural solutions that allow an agent to support persistent assumptions and simultaneously avoid inconsistencies across the hierarchical context that can lead to irrational behavior. Before introducing two new architectural solutions, however, we examine knowledge-based approaches and their consequences in order to provide further rationale for the architectural approach.

### 3.1 Knowledge-based Solutions

Inconsistency can be avoided in hierarchical agents by creating domain knowledge that recognizes potential inconsistencies and responds by removing assumptions. Many planning and agent systems use explicit domain knowledge to represent knowledge about the interactions among assertions in the world. For example, the Entropy Reduction Engine (ERE) (Bresina et al., 1993) is one agent system that relies on this *knowledge-based assumption consistency* (KBAC). ERE requires *domain constraints*, or knowledge that describes the physics of the task domain. Domain constraints identify impossible conditions. For instance, a domain constraint would indicate that a robot cannot occupy two different physical locations simultaneously.

In ERE, domain constraints are specifically used "to maintain consistency in the current world model state during execution" (Bresina et al., 1993, pp. 166). However, many other architectures use KBAC as well (perhaps in conjunction with other methods). KBAC knowledge can be viewed simply as domain knowledge that must be added to the system to achieve consistent behavior.

KBAC will always be necessary to maintain consistency among the assumptions within a level of the hierarchy. However, in order to guarantee consistency for assumptions distributed throughout the hierarchy, all possible interactions leading to inconsistency must be identified *throughout the hierarchy*. This knowledge engineering problem can add significant cost to agent development. A knowledge designer must not only specify the conditions under which an assumption is asserted but also *all* the conditions under which it must be removed. In the TacAir-Soar interception example, when the enemy plane flees, the agent requires knowledge that disables all assumptions that depend upon the number of enemy airplanes. Similarly, in the Dynamic Blocks World, the agent must have knowledge that recognizes any situation, in any subtask, that should cause the disabling of `empty`. In both cases, this KBAC knowledge "crosses" levels of the hierarchy. A complete KBAC solution requires that an agent's knowledge capture *all* potential dependencies between assumptions in a local subtask and any higher levels in the hierarchy.

Although it will be possible to encode complete across-level consistency knowledge for simple domains, experience in TacAir-Soar and other complex agent systems has convinced us that KBAC requires significant investments of time and energy. Further, because it is often not possible to enumerate all conditions under which an assumption must be removed, agents are also brittle, failing in difficult-to-understand, difficult-to-duplicate ways.

The insufficiency of knowledge-based solutions led us to consider architectural solutions to the problem. Architectural solutions eliminate the need for domain knowledge encoded only to address inconsistency between the hierarchical context and assumptions within a subtask. Thus, the cost of developing individual agents should be reduced. In addition to their generality, by definition, architectural solutions are also complete, and thus able to guarantee consistency between the hierarchical context and assumptions within a subtask, at all times, for any agent task. Such completeness should improve the robustness of agent systems, especially in situations not explicitly anticipated by their designers.

## 3.2 Assumption Justification

One potential architectural solution to the inconsistency problem is to justify each assumption in the hierarchy with respect to assertions in higher levels of the hierarchy. *Assumption Justification* is an extension of the truth maintenance approaches to consistency outlined previously. Each assumption in the hierarchy is treated as if it were an entailment with respect to dependent assertions higher in the hierarchy. A new data structure, the assumption justification, is created that captures the reasons in the hierarchical context for a particular assumption. Locally, an assumption is treated exactly like an assumption in a non-hierarchical system. However, when the assumption justification is no longer supported (indicating a change in the dependent hierarchical context), the architecture retracts the assumption.

Refer again to Figure 2. When the agent asserts $a_{34}$, the architecture builds an assumption justification for the assumption that includes $a_{22}$ and $a_{21}$. If the agent retracts $a_{21}$, the assumption justification for $a_{34}$ is no longer supported and the architecture also retracts $a_{34}$. The architecture ensures reasoning consistency across hierarchy levels because an assumption persists no longer than the context assertions that led to its creation.

Assumption Justification solves the inconsistency problem because all dependencies in the hierarchical context are captured in the justification. Within the subtask, domain knowledge is still required to ensure consistency among the enabled assumptions in the subtask. However, no across-level consistency knowledge is needed. Assumption Justification still supports local nonmonotonic and hypothetical reasoning. Thus, Assumption Justification appears to meet functional evaluation criteria. However, in order to assess its impact on performance, some implementation details must be considered.

### 3.2.1 Implementing Assumption Justification

Creating assumption justifications requires computing context dependencies for each assumption, similar to the computation of justifications for entailments.[2] Figure 7 outlines a procedure for computing the assumption justification data structure. This procedure is invoked when any assertion is created. This procedure creates assumption justifications for every assertion in a local subtask; that is, for entailments as well as assumptions. This approach allows the architecture to cache context dependencies for each local assertion. The advantage of this caching is that the architecture can simply concatenate the assumption justifications of the local assertions contributing directly to the creation of the assumption

---

2. The Assumption Justification procedure, as presented, requires that the inference engine record a justification for every assertion during the course of processing. In Soar, the architecture in which Assumption Justification was implemented, these calculations are available from its production rule matcher for both assumptions and entailments. However, justification calculations for assumptions may not be supported in other architectures, requiring modifications to the underlying inference engine. Laird and Rosenbloom (1995) and the *Soar User's Manual* (Laird, Congdon, & Coulter, 1999) describe the specific mechanisms of justification creation in Soar.

**PROC** *create_new_assertion*(. . .)
An assumption justification is computed when each new assertion A is created. Thus, assumption justifications are computed for both assumptions and entailments.
. . .
$A_{just} \leftarrow create\_justification(. . .)$
Justifications can be created via well-known, textbook algorithms (e.g., Forbus & deKleer, 1993; Russell & Norvig, 1995)
$A_{aj} \leftarrow make\_assumption\_justification\_for\_assertion(A)$
. . .
**END**

**PROC** *make_assumption_justification_for_assertion*(*assertion A*)
AJ ← NIL
**FOR** Each assertion $j$ in $A_{just}$, the justification of A
①　**IF** ($Level(j)$　closer to the root than　$Level(A)$)
AJ ← $append(j, AJ)$　(add $j$ to the assumption justification)
②　**ELSE**　($j$ and $A$ at the same level)
AJ ← $concatenate(j_{aj}, AJ)$　(add assumption justification of $j$ to assumption justification of A)

return AJ, a list of assertions comprising the assumption justification of A
**END**

**PROC** *Level*(*assertion　A*)
Return the subtask level associated with assertion A

Figure 7: A procedure for building assumption justifications.

(in ②). We chose this caching option for computing assumption justifications over an "on-demand" implementation that would, when an assumption was created, recursively follow local dependencies until all context dependencies were determined. The advantage of the caching implementation is that the context dependencies for any assertion must be computed only once, even when a local assertion contributes to the creation of multiple local assumptions.

The procedure that creates an assumption justification loops over the assertions in the justification of a new assertion **A**. The assertions in the justification can be either context or local assertions. Context assertions, in ①, are added to the assumption justification directly. However, local assertions should not be added to the assumption justification because the assumption justification should include only context dependencies. For example, the architecture can retract a local assertion for reasons other than a change in the hierarchical context (e.g., a non-monotonic reasoning step or a change in enabled assumptions in the subtask) and in these cases, the agent should not necessarily retract dependent assumptions. Because assumption justifications for all local assertions in the justification have already been computed (i.e., they are cached, as described above), the assumption justification of a local assertion $j$ can simply be added to the assumption justification of **A**, as in ②. For an on-demand implementation, the procedure here would recur through local assertions, until all context dependencies for local assertions contributing to **A** had been identified.

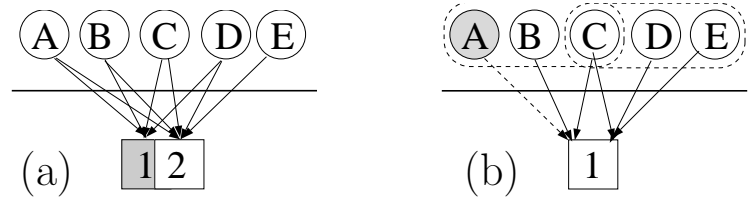

Figure 8: Technical problems with Assumption Justification. In (a), an assumption replaces another assumption nonmonotonically. In (b), multiple assumption justifications for the same assumption must be supported.

The worst-case computational complexity of this algorithm is polynomial in the number of assertions in the subtask. The addition of a higher-level assertion can be done in constant-time (a single pointer reference). However, ② must uniquely add context assertions from the assumption justification of the local assertion. There are at most $(n-1)$ local assumption justifications whenever the $nth$ assertion is created. Thus, the concatenation needs to be performed no more than $(n-1)$ times for any call to the assumption justification procedure. This limit provides an upper bound of $O(n)$ on the complexity of the assumption justification procedure: the worst-case cost of building an individual assumption justification is linear in the number of assertions, $n$, in the level. However, the architecture executes the assumption justification procedure for every assertion in the level. Thus, the worst case cost for building all the justifications in a particular level is $O(1 + 2 \ldots + n)$ or $O(n^2)$.

Non-monotonic changes complicate the implementation. The architecture must disable a replaced assumption, rather than delete it, because the initial assumption may need to be restored. For example, in Figure 8 (a), assume that the assertion of **E** leads to both the assertion of **2** in the local subtask and the retraction of **1** (i.e., **2** is a revision of **1**). If the agent retracts **E**, Assumption Justification will retract **2**, as desired, but it must also re-enable **1**. Thus, assumption **1** must remain available in memory, although disabled.

Figure 8 (b) illustrates a second problem. An assumption can have multiple assumption justifications. These justifications can change as reasoning progresses. Assumption **1** initially depends on assertions **A**, **B**, and **C** in higher levels. Now assume that later in the processing, the agent removes **A**, which normally would result in the retraction of **1**. However, in the meantime, the context has changed such that **1** is now also justified by {**C**, **D**, **E**}. Now when the agent removes **A**, the architecture should not immediately retract **1** but must determine if **1** is justified from other sources.

An implementation of Assumption Justification in Soar was completed by members of the Soar research group at the University of Michigan. Experiments using Air-Soar, a flight simulator domain (Pearson et al., 1993), showed that the overhead of maintaining all prior assumptions in a level produced a significant negative impact on agent performance. In this domain, Assumption Justification incurred significant computational cost, requiring at least 100% more time than the original Air-Soar agent. Further, the number of assumption justifications maintained within a level continued to grow during execution, for the reasons explained above. Some subtasks required minutes to execute as the aircraft performed a maneuver, leading to large (and problematic) increases in the amount of memory required. Thus, Assumption Justification failed to meet efficiency requirements on both theoretical and empirical grounds. Although the limitations of Assumption Justification might be improved by developing solutions to its technical problems, we abandoned further exploration of this approach after such strongly discouraging results.

### 3.3 Dynamic Hierarchical Justification

Figure 2 introduced the notion of a support set for subtasks. Both the Procedural Reasoning System (PRS) (Georgeff & Lansky, 1987) and Soar (Laird et al., 1987) use architectural mechanisms to retract complete levels of a subtask hierarchy when the support set no longer holds. In this section, we consider a solution that leverages the hierarchy maintenance function to ensure consistency between assumptions and the higher level context.

A significant disadvantage of the support set in existing systems is that it is *fixed*. In Soar, the support set is computed for the initiation of the subtask but is not updated to reflect reasoning that occurs within the subtask. For example, in Figure 2, suppose that assumption $a_{34}$ depends on assumptions $a_{22}$ and $a_{21}$ (represented by the dashed, arrowed lines). The support set does not include $a_{21}$; this assertion may not have even been present when $Subtask_3$ was created. When a local assumption depends on an assertion not in the support set, then a change in that assertion will not directly lead to the retraction of the assumption (or the subtask). Thus, approaches using such *Fixed Hierarchical Justification* (FHJ) still require knowledge-based solutions for consistency. FHJ is discussed further in Section 5.1.2.

We propose a novel solution, *Dynamic Hierarchical Justification* (DHJ), that is similar to Fixed Hierarchical Justification, but dynamically updates the support set as reasoning progresses. Assumption justifications for individual assumptions are unnecessary. However, one consequence of this simplification is that a subtask (and all assertions within it) will be retracted when the dependent context changes. Refer to Figure 2. When a DHJ agent asserts $a_{34}$ in Figure 2, the architecture updates the support set for $Subtask_3$ to include $a_{21}$. Assumption $a_{22}$ is already a member of the support set and need not be added again. When any member of the support set for $Subtask_3$ changes, the architecture retracts the entire subtask. Thus Dynamic Hierarchical Justification enforces reasoning consistency across the hierarchy because a *subtask* persists only as long as all dependent context assertions.

#### 3.3.1 Implementing Dynamic Hierarchical Justification

Figure 9 outlines the procedure for computing the support set in DHJ. As in Assumption Justification, the architecture can directly add context assertions to the support set ①. When the architecture computes the dependencies for a local assertion ③, the assertion is marked as having been inspected ④. Inspected assertions can simply be ignored in the future ②, because the architecture has already added the assertion's dependencies to the support set. The architecture will also ignore dependent, local assumptions ② because the dependencies of those assumptions will have already been added to the support set.

---

**PROC** *create_new_assertion*(...)
    Whenever a new assumption is asserted, the support set is updated
    to include any additional context dependencies.
    ...
    $A_{just} \leftarrow create\_justification(\ldots)$
    **IF** A is an assumption
        S is the subtask in which A is asserted
        $S_{support\_set} \leftarrow append(S_{support\_set}, add\_dependencies\_to\_support\_set(A))$
    ...
**END**

**PROC** *add_dependencies_to_support_set*(*assertion A*)
    **FOR** Each assertion $j$ in $A_{just}$, the justification of A
①     **IF** $\{Level(j)$    closer to the root than    $Level(A)\}$
            $append(j, S)$    (append context dependency to support set)

②     **ELSEIF** $\{Level(j)$    same as $Level(A)$    AND
             $j$ is NOT an assumption     AND
             $j$ has not previously been inspected $\}$
③         $S \leftarrow append(S, add\_dependencies\_to\_support\_set(j))$
             (compute support set dependencies for $j$ and add to S)
④         $j_{inspected} \leftarrow true$
             ($j$'s context dependencies have now been added to the support set)

    return S, the list of new dependencies in the support set
**END**

**PROC** *Level*(*assertion*    *A*)
    Return the subtask level associated with assertion A

---

Figure 9: A procedure for Dynamic Hierarchical Justification.

Because DHJ needs to inspect any local assertion only once, context dependencies are computed on-demand, rather than cached as in Assumption Justification. Condition ② will be true whenever there is a local entailment whose context dependencies have not yet been computed. These dependencies are determined by calling *add_dependencies_to_support_set* recursively. Recursive instantiations of *add_dependencies_to_support_set* each receive a local assertion in the justification of an uninspected entailment, $j$, and return a list comprising the context dependencies of $j$. The return value is then appended to the support set **S** in the prior instantiation of *add_dependencies_to_support_set*.

The recursive call to *add_dependencies_to_support_set* at ③ is the only non-constant time operation in the procedure. It must be made only once for any assertion $j_i$ and thus the worst case complexity to compute the dependencies is linear in the number of assertions in the level, as in Assumption Justification. However, DHJ requires only a single inspection of any individual assertion, rather than repeated inspections for each new assumption as in As-sumption Justification. Thus the architecture needs to call *add_dependencies_to_support_-set* at most $n$ times for any subtask consisting of $n$ assertions, and the worst case cost of updating the support set in a level remains $O(n)$. This reduction in complexity potentially makes Dynamic Hierarchical Justification a more efficient solution than Assumption Justification, especially as the number of local assertions increases.

Additionally, the two technical problems outlined for Assumption Justification do not impact DHJ. DHJ never needs to restore a previous assumption. When a dependency changes, the architecture retracts the entire level. Thus, DHJ can immediately delete replaced assumptions from memory. DHJ collects all dependencies for assumptions, so there is no need to switch from one justification to another. In Figure 8 (b), dependencies **A**, **B**, **C**, **D**, and **E** are all added to the support set. These simplifications can make the support set overly specific but reduce the memory and computation overhead required by Dynamic Hierarchical Justification.

DHJ retractions will sometimes be followed by the *regeneration* of the subtask and the re-assertion of reasoning that was retracted. For example, if the enemy plane fled as described in the TacAir-Soar scenario, DHJ would retract the entire level associated with the counting subtask. The count would then need to be re-started from the beginning. Section 3.4.3 examines potential problems introduced by interruption and regeneration. The cost incurred through the regeneration of previously-derived assertions is the primary drawback of Dynamic Hierarchical Justification.

## 3.4 Implications of Dynamic Hierarchical Justification

Dynamic Hierarchical Justification solves the specific problem of maintaining reasoning consistency in a hierarchy, guaranteeing consistency and utilizing an efficient algorithm. The heuristic DHJ employs assumes that assumptions are so closely associated with their subtasks that retracting subtasks is nearly equivalent to retracting individual assumptions. This section explores the implications of this heuristic, focusing on task decompositions, the impact on the agent's ability to use persistent assumptions, and the feasibility of interrupting an agent (with a subtask retraction) in the midst of reasoning.

### 3.4.1 The Influence of the Task Decomposition

An agent's reasoning can be viewed as knowledge search (Newell, 1990). From this perspective, the inconsistency problem is failure to backtrack in knowledge search. The world changes, leading to changes in the agent hierarchy. The agent must retract some of the knowledge it has previously asserted, so it should backtrack to a knowledge state consistent with the world state.[3] Each solution can be described in terms of the way its achieves (or avoids) backtracking in the knowledge search. For instance, KBAC leads to knowledge-based backtracking, in which the KBAC knowledge tells the agent how to correct its assumptions given the current situation.

---

3. Obviously, this world state will usually be different than the agent's initial state and it is often impossible to return to a prior state in an execution system. We use "backtrack" in this section to refer to the retraction of asserted execution knowledge such that any remaining asserted knowledge is consistent with the currently perceived world state.

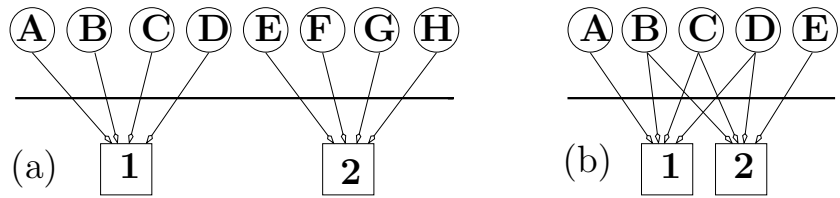

Figure 10: Examples of (a) disjoint dependencies and (b) intersecting assumption dependencies.

Assumption Justification is a form of dependency-directed backtracking (Stallman & Sussman, 1977). In dependency-directed backtracking, regardless of the chronological order in which an architecture makes assertions, the architecture can identify and retract those assertions that contributed to failure in a search and retain all other assertions. In Assumption Justification, the architecture retracts only those assumptions that are directly affected by a change in the context. Assumptions created later in the processing, not dependent on the change, are unaffected. Consider the examples in Figure 10. In (a), assumptions **1** and **2** each depend upon disjoint sets of assertions. With Assumption Justification, removal of any assertion in **1**'s assumption justification will result in the retraction of **1**; **2** is unchanged, even if the architecture asserted **2** after **1**.

Dynamic Hierarchical Justification is similar to *backjumping* (Gaschnig, 1979). Backjumping heuristically determines the state to which a current search should backtrack or "backjump." The heuristics used by backjumping are based on syntactic features of the problem. For instance, in constraint satisfaction problems, the backjumping algorithm identifies which variable assignments are related to other variable assignments via the constraints specified in the problem definition. When a violation is discovered, the algorithm backtracks to the most recent, related variable (Dechter, 1990). Intervening variable assignments are discarded. In DHJ, when an assertion in the hierarchy changes, the system "backjumps" in its knowledge search to the highest subtask in the hierarchy not dependent on the change. In Figure 10 (a) all dependent assertions are collected in the support set for the subtask. If any of the higher level assertions change, the entire subtask is removed.

When using DHJ, as in backjumping, some previous knowledge search may need to be repeated after backtracking. Assume the removal of the subtask in Figure 10 (a) was due to a change in **A**. If a similar subtask is reinitiated, assumption **2** may need to be regenerated. This regeneration is *unnecessary* because **2** did not need to be retracted to avoid inconsistency. Under Dynamic Hierarchical Justification, the agent retracts *all* reasoning in the dependent subtask (and all lower levels in the hierarchy); assertions not dependent on the change in the context can also be removed. Thus, like backjumping, DHJ uses a syntactic feature of reasoning (decomposition into subtasks) to choose a backtracking point and this backtracking is not always as conservative as possible.

Although the subtask decomposition is a syntactic feature of the knowledge search, it is a strongly principled one, reflecting a semantic analysis of the task by a knowledge designer. Hierarchical task decomposition is based on the premise that tasks can be broken down into discrete units that have little interaction with other units; they are *nearly decom-* *posable* (Simon, 1969). Thus, the goal of a hierarchical decomposition is to separate mostly independent subtasks from one another. A consequence of this separation is that dependencies in higher levels will be limited as much as possible (interaction between subtasks should be minimized) and the dependencies among the assertions in any particular subtask will be shared (otherwise, the subtask could be subdivided into two or more independent subtasks). Of course, it is often possible to decompose a given task in many different ways. In most cases the domain imposes minimal constraint and the knowledge engineer has significant latitude in crafting the task decomposition.

In the situation illustrated in Figure 10, (b) would be a more complete decomposition of a task by the knowledge engineer than (a), assuming the two alternatives represent a decomposition of the same task. In (b), the number of dependent assertions does not necessarily grow as a function of the number of assumptions in the local level, while in (a) it does. Further, in (a), two independent assumptions are being pursued. These assumptions could potentially be inferred in separate subtasks in an alternate decomposition. In (b), on the other hand, the assumptions in the subtask are closely tied together in terms of their dependencies and thus better asserted within the same subtask. Because the dependencies of assumptions **1** and **2** have considerable overlap in (b), Assumption Justification pays a high overhead cost to track individual assumptions because (most) everything in the local subtask would be removed simultaneously if assertions **B**, **C**, or **D** changed. Because DHJ incurs no such overhead, DHJ is a better choice when the intersection between assumption dependencies is high. Task knowledge structured more like the situation in (b), rather than (a) would lead to few *unnecessary* retractions. Because (b) appears to better reflect well-decomposed tasks, Dynamic Hierarchical Justification will constrain the knowledge development process and improve the resulting decompositions. Consequently, nearly-decomposed tasks should allow DHJ to avoid most unnecessary regenerations while avoiding the processing overhead of Assumption Justification.

### 3.4.2 Limiting Persistence under DHJ

DHJ limits persistence in subtasks, resulting in assumptions that are not as persistent as assumptions in typical truth maintenance systems. This section explores the consequences of these limitations to determine if DHJ architectures[4] can still provide the persistence necessary for agent execution (Section 2.1).

DHJ will retract a subtask when a potential inconsistency could impact hypothetical and recursive reasoning like counting. Consider the aircraft classification and counting example. Perhaps an aircraft's altitude contributes to a hypothetical classification of the aircraft (e.g., particular altitude and speed combinations might suggest a reconnaissance aircraft). The agent would create assumptions locally that depend on this aircraft's altitude. If the altitude (or an altitude boundary) changes, then the assumption should be retracted. This retraction is required to avoid inconsistency. If the contact's altitude no longer suggests that it is a reconnaissance aircraft, then the assumption that depended on that assertion should be removed. DHJ captures these dependencies and performs the retraction. The agent now has the opportunity to reconsider the classification of the aircraft, or pursue other tasks if the classification is no longer important.

---

4. For clarity, and because Assumption Justification has already been eliminated as a candidate solution, the following discussion focuses exclusively on DHJ. However, Assumption Justification limits persistence similarly.

DHJ will also retract a subtask if an assumption was created in the local subtask for the purpose of remembering some input (or elaboration of input). For example, if an agent needed to remember a particular aircraft's altitude at a particular point in time, then that assumption cannot be stored in a local subtask. DHJ limits persistence in such a way that remembering *within* a local subtask is generally impossible.

In order to remember previous situations, assumptions can be asserted in the root task. Any assumption asserted in this level will never be retracted because there are no higher level dependencies (assuming percepts are associated with the top level and not a higher "input level," as in Theo, Mitchell et al., 1991). The primary drawback of this requirement for remembering is that remembered items are no longer local to the subtask that created them, requiring additional domain knowledge to manage remembered assumptions. However, remembering already requires domain knowledge; it is not possible to remember an assertion regardless of its dependencies and also be able to retract it architecturally.

These examples show that Dynamic Hierarchical Justification still allows all forms of persistence, but trades capturing dependencies for nonmonotonic assumptions in local subtasks with remembering assumptions in the root task, where no dependencies are captured. Because DHJ forces remembered items into the root task, it also suggests that a fundamental aspect of this root task should be managing these remembered assumptions. We view this requirement as a positive consequence of DHJ, because it forces knowledge engineers to better recognize the reasons for creating an assumption (e.g., remembering vs. a hypothetical) and circumscribes remembering so that we can now develop or adopt functional or temporal theories to manage assumptions created for remembering (e.g., Allen, 1991; Altmann & Gray, 2002).

### 3.4.3 Recovery from Interruption with DHJ

Dynamic Hierarchical Justification makes an agent more reactive to its environment, ensuring that relevant changes in the environment lead to the retraction of any dependent subtasks. DHJ imposes an automatic interruption of the agent for a subtask retraction, without evaluating the state of the system first. Although automatic interruption increases the reactivity of the system, it can lead to difficulties if there is no way to override it. In this section we examine two cases where uncontrolled interruption can cause problems. The problems arise because DHJ biases the system to be reactive; that is, to respond automatically to changes in the environment without deliberation. However, in both cases, additional agent knowledge can overcome that bias and make the system more deliberate and avoid uncontrolled interruption.

The first problem arises when there is a sequence of actions that must be completed without interruption in order for a subgoal to be achieved. If the processing is interrupted, then it is possible, because of the dynamics of the world, that the task cannot be resumed. For example, imagine an aircraft nearing the point where it can launch a missile at a target. When the task is interrupted and then resumed, the aircraft's position may have changed enough, relative to the target, that additional steering commands are necessary before the missile can be launched. In this case, it may be preferable not to interrupt the original launch sequence once it has begun.

Consider two possible approaches to achieving this capability with Dynamic Hierarchical Justification architectures. The first is to move the processing to the root task. Because the root task is not interrupted, the processing will not be interrupted. However, this approach greatly restricts how a task can be hierarchically decomposed and thus should be considered only as a last resort. The second approach is to add new reasoning for the task that "freezes" the external situation with respect to additional reasoning in the subtask. The new processing initiates the execution of the subtask and creates persistent structures in the root task. These persistent structures represent a deliberate commitment to not being interrupted. The remaining processing in the subtask accesses only these structures in the execution of the task. Thus, because they are persistent, even if there are changes to the surrounding situation that would have interrupted the subtask, its processing is now insensitive to those changes and interruption is prevented. This approach also requires additional reasoning to recognize completion of the uninterruptible behavior and remove the persistent structures built by the initial subtask. This reasoning reflects a deliberate act, signaling that the commitment no longer holds. In the abstract, together these additions provide a mechanism for overcoming automatic interruption. The disadvantage of this approach is that, as part of the system design, those subgoals that cannot be interrupted must be identified beforehand. For those subtasks, additional agent knowledge must be implemented to create and remove encapsulations of any dynamic data.

A more critical problem for DHJ is the "Wesson Oil" problem: when someone is cooking dinner and a higher-priority activity suddenly occurs (a hurt child), the cook should turn off the stove (a "cleanup" procedure) before leaving for the hospital (Gat, 1991b). This problem occurs when there is a change in the hierarchical context at a level far from the terminal level of the hierarchy. In this situation, similar tasks may not be resumed or initiated following the interruption. The agent must therefore recognize whether "cleanup" of the external and/or internal states is necessary, and, if so, perform that cleanup. Even with DHJ, the agent can still behave appropriately if it has the right knowledge. In particular, the agent must be able to recognize partially completed tasks (like cooking dinner) and be able to select cleanup actions specific to the task state (like turning off a stove burner). Because DHJ requires all remembered assumptions to be asserted in the root level of the hierarchy, this recognition task has internal state available; it need not try to reconstruct that state from the external environment alone. However, it does require some analysis of the task domain(s) by a knowledge engineer so that any interruptible activity requiring cleanup include triggering assertions for cleanup in the root task.

This work was prompted by a desire for architectural solutions to inconsistency, yet maintaining consistency efficiently can lead to interruptions, which, under DHJ, requires knowledge-based solutions to problems arising from automatic interruption.[5] However, most of the requirements imposed by DHJ are positive consequences. Subtask retractions and observed recovery in the development process help define what must be remembered in the root task for cleanup, which is significantly different than the laborious process of debugging agent programs that are failing due to inconsistency. In theory, Dynamic Hierarchical Justification imposes requirements for handling interruptions that do pose serious questions about its overall utility. In practice, we have not found addressing these questions to be a problem in a variety of recent agent implementations using the Soar-DHJ architecture (e.g., Laird, 2001; Wray et al., 2002).

---

5. Dynamic Hierarchical Justification could also be used as a trigger for meta-level deliberation rather than immediate subtask retraction. It would then possibly provide an architectural solution to the question of when to deliberate about potential inconsistency for *intention reconsideration* (see Section 5.2).

## 4. Empirical Evaluation of Dynamic Hierarchical Justification

Architectural mechanisms like DHJ must be efficient. We have demonstrated that the algorithm itself is efficient, but the question of its impact on the overall behavior generation capability of an agent remains an open question due to interruption and regeneration. Given the complexity of both agent-based systems and the domains in which they are applied, analytical evaluations must be extremely narrow in scope, and even then require specialized techniques (Wooldridge, 2000). This section instead pursues an empirical evaluation of Dynamic Hierarchical Justification, focusing on efficiency and responsiveness in two domains at extremes in the continua of agent domain characteristics. Because the architectural solution to inconsistency was motivated by the cost (and incompleteness) of knowledge-based solutions, knowledge development costs will also be estimated.

### 4.1 Methodological Issues

Dynamic Hierarchical Justification is a general solution, applicable in a wide range of agent tasks. In order to evaluate such a solution, a number of methodological issues must be addressed. The following describes three important issues and the choices made for this evaluation.

#### 4.1.1 RELATIVE *vs.* ABSOLUTE EVALUATION

What constitutes "good" or "poor" cost and performance evaluations? In general, an absolute evaluation of performance and cost is difficult because the task itself, in addition to the agent's knowledge and architecture, determines overall cost and performance results.

We circumvent this problem by making relative comparisons between agents using the original, Fixed Hierarchical Justification Soar architecture ("FHJ agents") and new agents ("DHJ agents"). The FHJ agents provide cost and performance benchmarks, obviating the need for absolute evaluations.

#### 4.1.2 ADDRESSING MULTIPLE DEGREES OF FREEDOM IN AGENT DESIGN

Even when architecture and task are fixed, many different functional agents can be developed. How can one know if comparative results are valid and general if the experimenter has control over both benchmarks and new agents?

DHJ agents will be compared to agents previously implemented by others. Such systems will provide good performance targets, because they were optimized for performance, and will minimize bias, because they were developed independently.

FHJ systems were used as fixed benchmarks, and were not modified. DHJ agents use the identical task decompositions employed by the FHJ agents and the same initial knowledge base. We observed opportunities to improve performance in the DHJ agents by modifying either the task decomposition or re-designing significant portions of the agent knowledge base. However, agent knowledge was modified only when necessary for correct behavior, in order to ensure that DHJ agents remained tightly constrained by their FHJ counterparts, thus limiting bias in the evaluation.

### 4.1.3 The Choice of Representative Tasks

This evaluation will be limited to execution agents in the Dynamic Blocks World and in a reduced-knowledge version of TacAir-Soar ("micro-TacAir-Soar"). The choice of only a few tasks or domains is a considerable drawback of benchmarks (Hanks, Pollack, & Cohen, 1993). Although these choices were motivated primarily by the availability of domains with pre-existing FHJ agents, the two domains do represent opposite extremes for many domain characteristics. Micro-TacAir-Soar, like TacAir-Soar, is inaccessible, nondeterministic, dynamic, and continuous, while the Dynamic Blocks World simulator used in the experiments is accessible, deterministic, static and discrete. The primary motivation for using the Dynamic Blocks World, which is less representative of typical agent tasks than Micro-TacAir-Soar, is to assess the cost of employing DHJ in a domain where *a priori* it appears it would not be useful (although Section 6 suggests DHJ can prove useful even in relatively static domains). Thus, the Dynamic Blocks World will provide a baseline for the actual cost of deploying the algorithm, even though little benefit is expected from its deployment in this domain.

## 4.2 Evaluation Hypotheses

Although specific expectations will differ in different domains, differences in the dimensions of knowledge cost and performance can be anticipated when comparing DHJ agents to baseline agents. The following discusses the expectations and the metric(s) used for each dimension.

### 4.2.1 Knowledge Engineering Cost

Knowledge engineering effort in DHJ agents should decrease in comparison to previously developed agents. Knowledge in Soar is represented with production rules. Each production represents a single, independent knowledge unit. We assume the addition of more productions represents an increase in cost and measure knowledge cost by counting the number of productions in each type of agent. The number of productions, of course, provides only a coarse metric of cost because the complexity of individual productions varies significantly. However, the productions that will be removed in DHJ agents are often the most difficult ones to create. Therefore, the difference in number of productions is probably a conservative metric for knowledge cost in DHJ.

### 4.2.2 Performance: Efficiency and Responsiveness

In general, overall performance should change little in DHJ agents, as compared to FHJ counterparts. Although Dynamic Hierarchical Justification does add a new architectural mechanism, the algorithm itself is efficient and should not contribute to significant differences in performance. Further, less domain knowledge will need to be asserted because all across-level consistency knowledge is now incorporated in the architecture. Thus, if applying across-level KBAC knowledge represented a significant expense in the overall cost of executing a task, DHJ agents might perform better than FHJ agents.

There are two specific exceptions to this expectation. First, in domains where consistency knowledge is (mostly) unnecessary for task performance, FHJ agents may perform better than DHJ agents. For example, the Dynamic Blocks World requires little consistency knowledge but the DHJ architecture will still update the support set, even though few inconsistency-causing context changes should arise.

Second, if regeneration is problematic, overall performance will suffer. In DHJ, whenever the dependent context changes, a subtask will be retracted. If the change does not lead to a different choice of subtask, the subtask will be necessarily regenerated. Thus, under DHJ, some subtask regeneration will occur, and, if regeneration is significant, performance degradation will result.

CPU execution time provides a simple, single dimension of gross performance. The CPU time reported for individual experiments reflects the time the agent spends reasoning and initiating actions rather than the time it takes to execute those actions in the environment.
**Decisions:** In Soar, subtasks correspond to the selection of *operators* and subgoals for implementing operators. The selection of an operator is called a *decision*. When Soar selects an operator, it tries to apply the operator. Soar reaches an *impasse* when it cannot apply a newly selected operator. These non-primitive operators lead to the generation of a subgoal in the subsequent decision. For example, Soar selects the `put-down` operator in one decision and creates a subgoal to implement `put-down` in the subsequent decision. Together, these two steps constitute the notion of a subtask in Soar.

The number of decisions can thus be used as an indication of the number of subtasks undertaken for a task. In FHJ, a subtask was generally never interrupted until it terminated (either successfully or unsuccessfully). In DHJ, subtasks will be interrupted whenever a dependent change occurs. Thus, decisions should increase in DHJ agents because subtasks will be interrupted and re-started. Further, if decisions increase substantially (suggesting significant regeneration), overall performance will degrade.
**Production Firings:** A production rule "fires" when its conditions match and its result is applied to the current situation. Production firings should decrease in DHJ for two reasons. First, any across-level consistency knowledge that was previously used in FHJ agents will no longer be necessary (or represented); therefore, this knowledge will not be accessed. Second, any reasoning that occurred after inconsistency arose in FHJ agents will be interrupted and eliminated. However, production firings will increase if significant regeneration is necessary.

## 4.3 Empirical Evaluation in the Blocks World

Agents in this Dynamic Blocks World domain have execution knowledge to transform any initial configuration of three blocks into an ordered tower using a simulated gripper arm. The table in this simulation has a width of nine blocks. The agent's task goal is always to build the **1-on-2-on-3** tower. Each agent built a tower from each of the resulting 981 unique, non-goal, initial configurations of blocks. Table 1 summarizes the results of these tasks. As expected, total knowledge decreased. Overall performance improved. Decisions increased, as expected, but the number of rule firings increased as well, which was not anticipated.

|              | FHJ       |      | DHJ       |       |
|--------------|-----------|------|-----------|-------|
|              | $\bar{x}$ | s.d. | $\bar{x}$ | s.d.  |
| Rules            | 188   | —     | 175   | —     |
| Decision Avg.    | 87.1  | 20.9  | 141.1 | 38.7  |
| Avg. Rule Firings| 720.3 | 153.5 | 855.6 | 199.6 |
| Avg. CPU Time (ms)| 413.1 | 121.6 | 391.6 | 114.0 |

Table 1: Summary of knowledge and performance data from the Blocks World. The agents performed the tower-building task for each of 981 configurations. Task order was randomly determined.

### 4.3.1 Knowledge Differences

Total knowledge decreased about 7% in the DHJ agent. This small reduction is consistent with expectation. The aggregate comparison is misleading because knowledge was both added (16 productions) and deleted (29).

**Removing Consistency Knowledge**: In Soar, the subtask operator and subgoal are terminated separately. Soar monitors impasse-causing assertions to determine if a subgoal (such as the subtask goal) should be removed via FHJ. However, the removal of a subtask operator requires knowledge. The original, FHJ architecture treats the initiation of an operator as a persistent assumption and requires knowledge to recognize when a selected operator should be interrupted or terminated. This knowledge can be categorized as consistency knowledge because it determines the time at which a subtask should be terminated, even when the initiating conditions for the subtask no longer hold.

In DHJ, only the effects of operators are persistent; all other assertions are entailments of the situation. Thus, the initiation of a subtask is now treated as an entailment and a subtask remains selected only as long as the initiation conditions for the subtask hold. This change removes the need for knowledge to terminate the subtask: when the subtask initiation conditions are no longer true, the subtask is automatically retracted. Thus, termination knowledge was removed for all subtask operators.

**Filling Gaps in Domain Knowledge**: The persistence of subtasks in the original architecture allows FHJ agents to ignore large parts of the state space in their domain knowledge. For example, the knowledge that initiates `stack` and `put-on-table` subtasks assumes that the gripper is currently not holding a block. As these tasks are executed, the gripper, of course, does grasp individual blocks. The conditions for initiating `stack` or `put-on-table` when holding a block were ignored in the original domain knowledge.

The DHJ agent now requires knowledge to determine which subtasks it should choose when holding blocks, because subtasks can be interrupted while the agent still holds a block. 16 productions were necessary, primarily for the `stack` and `put-on-table` operators. It is important to note that this knowledge is necessary domain knowledge. FHJ agents could not solve any problem in which they began a task holding a block because they lacked domain knowledge for these states. These additions are thus a positive consequence of DHJ. The architecture's enforcement of consistency revealed gaps in domain knowledge.

### 4.3.2 PERFORMANCE DIFFERENCES

Somewhat surprisingly, overall performance of the DHJ agents (measured in CPU time) improves slightly in comparison to the FHJ agents, even though both decisions and production firings increase. Each of the Soar-specific performance metrics are considered individually below, and then the overall performance improvement is considered.

**Decisions**: FHJ agents, on average, made considerably fewer decisions than DHJ agents. The difference was consistent across every task. These additional decisions result from the removal and subsequent regeneration of subtasks. For example, when the agent picks up a block in pursuit of a `stack` task, the selection of the `stack` task must be regenerated. The knowledge in the DHJ agents could be modified to avoid testing specific configurations of blocks and thus avoid many of these regenerations.

**Production Firings**: The number of production firings also increased in the Blocks World. The increase in production firings can be attributed to the knowledge added to the system and the regeneration of subtasks that made the additions necessary. The relative increase in number of production firings (19%) was much smaller than the increase in decisions (62%). The smaller difference can be attributed to the productions that were removed (and thus did not fire).

**CPU Time**: Generally, when production firings increase in Soar, an increase in CPU time is expected. However, CPU time in DHJ decreased slightly in comparison to FHJ even though production firings increased. To explain this result, some additional aspects of Soar's processing must be considered.

The match cost of a production is not constant but grows linearly with the number of *tokens*, partial instantiations of the production (Tambe, 1991). Each token indicates what conditions in the production have matched and the variable bindings for those conditions. Thus, each token represents a node in a search over the agent's memory for matching instantiation(s) of the production. The more specific a production's conditions are, the more constrained the search through memory, and thus it costs less to generate the instantiation.

The new productions added to the DHJ Blocks World agent were more specific to the agent's memory (i.e., its external and internal state) than the productions removed. Further, simply having fewer total productions also can reduce the amount of total search in memory.[6] An informal inspection of the match time and tokens for several FHJ and DHJ runs showed that the number of tokens decreased in DHJ by 10-15%. This reduction in token activity is the primary source of improvement in Dynamic Blocks World DHJ agent CPU time. This improvement, of course, is not a general result and provides no guarantee that in some other task or domain the cost of matching will not increase rather than decrease.

---

6. The RETE algorithm (Forgy, 1979) shares condition elements across different productions. Thus, the removal of productions only decreases the total search if removed productions contain condition elements not appearing in the remaining productions. We did not perform an exhaustive analysis of the condition elements to determine if the removed productions reduce the number of unique condition elements in the RETE network.

## 4.4 Empirical Evaluation in $\mu$TacAir-Soar

Converting TacAir-Soar to the DHJ architecture would be very expensive, requiring many months of effort. DHJ agents were instead developed for a research and instruction version of TacAir-Soar, "Micro-TacAir-Soar" ($\mu$TAS). $\mu$TAS agents use the TacAir-Soar simulation environment (ModSAF) and interface but have knowledge to fly only a few missions, resulting in an order of magnitude decrease in the number of productions in the agents. However, $\mu$TAS uses the same tactics and doctrine for its missions as TacAir-Soar.

In $\mu$TAS, a team of two agents ("lead" and "wing") fly the patrol mission described previously. They engage any hostile aircraft that are headed toward them and are within a specific range. The lead agent's primary role is to fly the patrol route and intercept enemy planes. The wing's responsibility is to fly in formation with the lead. Because the total knowledge is significantly reduced, converting $\mu$TAS DHJ agents should be relatively inexpensive. However, the results should be representative of TacAir-Soar because $\mu$TAS retains the complexity and dynamics of TacAir-Soar.

The patrol mission has no clearly-defined task termination condition like the Dynamic Blocks World. To address this problem, each agent in the simulation executes for ten minutes of simulator time. During this time, each agent has the opportunity to take off, fly in formation with its partner on patrol, intercept one enemy agent, and return to patrol after the intercept. In an actual TacAir-Soar scenario, these activities would normally be separated by much larger time scales. However, an agent spends much of its time on a patrol mission simply monitoring the situation (waiting), rather than taking new actions. Ten minutes of simulated time proved to be brief enough that overall behavior was not dominated by wait-states, while also providing time for a natural flow of events.

When running for a fixed period of time, an increase in the number of decisions can be attributed to regeneration or simply an improvement in decision cycle time. We avoid this potential confusion by running the simulator with a constant cycle time. In this mode, each simulator update represents 67 milliseconds of simulated time. Because each agent now runs for a fixed period of time with fixed updates, each FHJ and DHJ agent will execute the same number of decisions. Any problems due to regeneration will be apparent in the number of rule firings and degradation in responsiveness. Additionally, the general results do not change significantly if the scenarios are executed with the real-time mode normally used for TacAir-Soar agents. The fixed cycle simply eliminates some variability.

Although the patrol scenario was designed to minimize variation from run to run, the $\mu$TAS simulator is inherently stochastic and the specific actions taken by an agent and the time course of those actions varies when the same task is repeated. To control for this variation, each scenario was run for the lead and wing agents approximately 50 times. Logging and data collection significantly impacted CPU time and other performance statistics. In order to control for this effect, we actually ran each scenario 99 times, randomly choosing one agent (lead or wing) to perform logging functions (and discarding its performance measures). The other agent performed no logging functions. Data from logging agents was used to create Figure 9. The performance measures of the "no logging" agents were recorded at the conclusion of each scenario and are summarized in Table 2.

| | Lead Agent | | | | Wing Agent | | | |
|---|---|---|---|---|---|---|---|---|
| | FHJ | | DHJ | | FHJ | | DHJ | |
| Rules | 591 | | 539 | | 591 | | 539 | |
| Number of runs ($n$) | 43 | | 53 | | 56 | | 46 | |
| | $\bar{x}$ | s.d. | $\bar{x}$ | s.d. | $\bar{x}$ | s.d. | $\bar{x}$ | s.d. |
| Decisions | 8974 | 0.0 | 8974 | 0.0 | 8958 | 0.0 | 8958 | 0.0 |
| Outputs | 109.1 | 6.71 | 142.8 | 7.03 | 1704 | 42.7 | 869 | 12.8 |
| Rule Firings | 2438 | 122 | 2064 | 81.1 | 16540 | 398 | 6321 | 104 |
| CPU Time (msec) | 1683 | 301 | 1030 | 242 | 12576 | 861 | 2175 | 389 |

Table 2: Summary of $\mu$TAS run data.

### 4.4.1 IMPROVING TASK DECOMPOSITIONS

$\mu$TacAir-Soar DHJ agents required extensive knowledge revision. Such revision was not unexpected. For instance, unlike the Dynamic Blocks World, $\mu$TAS agents remember many percepts, such as the last known location of an enemy aircraft. As previously described, assertions for remembering must now be located in the root level of the hierarchy, thus requiring some knowledge revision. However, other problems were discovered. In some cases, FHJ agents *took advantage* of inconsistency in asserted knowledge. In other words, the FHJ agent not only allowed inconsistency in the assertions but actually depended on those inconsistencies to apply new knowledge. There were two major categories of this knowledge. "Within-level" consistency knowledge recognized specific inconsistencies (e.g., retraction of the proposal for the subtask) as a trigger for actions such as "clean up" of the subtask state. "Complex subtasks" allowed the non-interruptible execution of a complex procedure regardless of the continuing acceptability of the subtask. In both cases, agent knowledge was modified to remove any dependence on inconsistency. Appendix A provides further explanation of the original knowledge and subsequent changes. Section 4.4.3 summarizes the changes quantitatively.

### 4.4.2 RESULTS

Table 2 lists average data for the FHJ and DHJ lead and wing agents for the patrol/intercept scenario after the modifications to the DHJ agent's knowledge base were completed. The results in this domain are consistent with expectations: total knowledge decreases, rule firings decrease and performance improves, substantially so for the DHJ wing agent. The following sections explore each of these results in greater detail.

### 4.4.3 KNOWLEDGE DIFFERENCES

Table 3 quantifies the changes to the Soar production rules described above.[7] Modifications include deletions, additions and changes. A rule was considered "changed" only if its conditions changed slightly, but it made the same type of computation for the same subtask. For example, most of the "changed" within-level consistency knowledge now refers to an entailed structure rather than one created as an assumption, but that structure is located in the same subtask. This somewhat restrictive definition of a change inflates the addition and deletion accounting. In many cases a production was "deleted" and then immediately "added" to a different subtask. For example, the productions that manipulate motor commands were all moved from local subtasks to the highest subtask. Almost all the additions and deletions in the "Remembering" category can be attributed to this move, which required no synthesis of new production knowledge.

---

7. The DHJ agent data was generated with a knowledge base that included some changes to accommodate learning (Wray, 1998) and these changes are included in the table for completeness. The presence of these rules in the knowledge base has negligible impact on the performance data reported here.

| | Across-level Consistency | Remembering | Within-level Consistency | Complex Subtasks | Learning | Miscellaneous | TOTALS |
|---|---|---|---|---|---|---|---|
| FHJ Agent: | | | | | | | 591 |
| Deletions: | 44 | 36 | 9 | 10 | 4 | 8 | (111) |
| Additions: | 0 | 32 | 5 | 21 | 0 | 1 | 59 |
| DHJ Agent: | | | | | | | 539 |
| Additional Changes: | 0 | 33 | 8 | 0 | 24 | 0 | 65 |

Table 3: Quantitative summary of changes to production rules in the FHJ agent knowledge base for DHJ agents.

Total knowledge required for the DHJ agents decreased. This approximately 9% reduction was achieved by making some type of modification to about 40% of the FHJ agent rules, and may seem a modest gain, given the conversion cost. However, this cost is an artifact of the chosen methodology. Had the DHJ agents been constructed in this domain without previously existing FHJ agents, at least a 9% decrease in the total knowledge would be expected. This result thus suggests a reduction in the cost of the agent design. The high conversion cost does suggest that converting a much larger system, like TacAir-Soar, would probably be very costly. On the other hand, the modifications were made evident by identifiable regenerations in the architecture. Thus, the 235 total changes made to the FHJ knowledge base were much easier to make than constructing a similar number of rules.

#### 4.4.4 Performance Differences

As the performance results in Table 2 show, DHJ agents improved in performance relative to their FHJ peers. However, the improvements of the lead and wing agents was substantially different. Differences in the tasks of lead and wing pilots led to the differences in relative improvements.

**Lead and Wing Agents**: The lead and wing agent share the same knowledge base but perform different tasks in the $\mu$TAS scenario.[8] These differences lead to differences in their absolute performance. Recall that the lead's primary responsibility is to fly the patrol route and intercept enemy aircraft. On the other hand, the wing's primary mission role is to follow the lead. These different tasks require different responses in the agents.

---

8. The agents share the same knowledge base because they can dynamically swap roles during execution. For instance, if the lead exhausts its long-range missiles, it will order the wing to take over the lead role, and then take the role of wing itself.

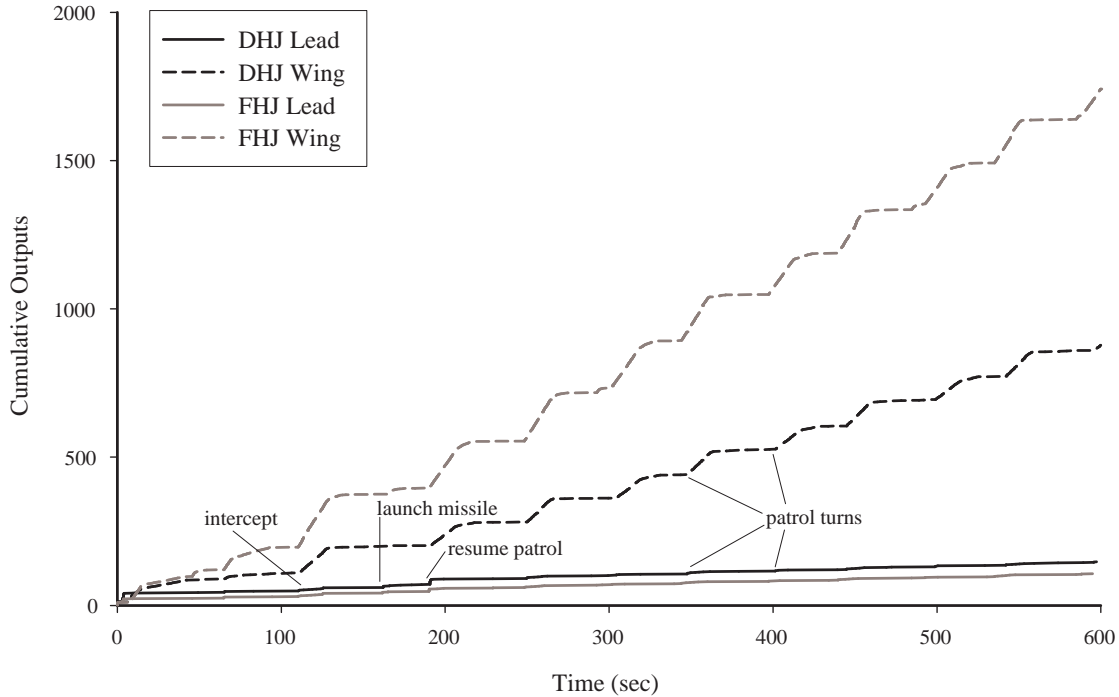

Figure 11: Cumulative outputs over the course of one ten minute scenario for DHJ (black) and FHJ (gray) agents. Cumulative outputs for lead agents are represented with solid lines, wing agents with dashed lines.

An agent's overall reasoning activity is often correlated with its output activity; that is, the commands it sends to the external environment to take action in it. Figure 11 summarizes the output activity of two pairs of lead and wing agents (FHJ & DHJ) over the course of a ten-minute scenario. The output activity of both leads is mostly concentrated at a few places over the course of the scenario (`take-off`, `intercept`, `launch-missile`, and when resuming patrol following the intercept). The wings' most concentrated output activity occurs when the leads turn to a new leg of the patrol and the wings must follow the lead through a 180 degree turn. In the remainder of this section, we focus only on DHJ agents to contrast lead and wing agent behavior. The discussion of the performance metrics will examine differences between FHJ and DHJ leads and wings.

The lead actually spends most of the scenario waiting, with short bursts of reasoning and output activity occurring at tactically important junctures in the scenario. On patrol, the lead flies straight and makes a decision to turn when it reaches the end of a patrol leg. The lead monitors the environment and searches for enemy planes. This search is (mostly) passive; the agent's radar notifies the agent if any new entities have been detected. After detecting and classifying an enemy plane as a potential threat, the lead commits to an intercept. The lead immediately makes a number of course, speed, and altitude adjustments, based on the tactical situation. These actions are evident in the figure by the pulse labeled "intercept." The lead spends most of the time in the intercept closing the distance between the aircraft to get within weapon range, again having to maneuver very little and thus requiring few actions in the environment (thus the relatively flat slope following intercept). When the agent reaches missile range of the enemy plane, the lead executes a number of actions very quickly. The lead steers the plane into a launch window for the missile, pushes the fire button, waits for the missile to clear, and then determines a course to maintain radar contact as the missile flies to its target (at `launch-missile`). Once the intercept has been completed, the lead resumes its patrol task. Again, it issues a large number of output commands in a short period of time. These examples show that the lead's reasoning focuses primarily on reacting to discrete changes in the tactical situation (patrol leg ended, enemy in range, etc.) and the behavior generally requires little continuous adjustment.

The execution of the wing's `follow-leader` task, on the other hand, requires reaction to continuous change in the lead's position in order to maintain formation. Position corrections require observing the lead's position, recognizing an undesired separation in the formation, and then responding by adjusting speed, course, altitude, etc. Because the wing is following the lead throughout the scenario, it is executing this position maintenance knowledge almost constantly. When the lead is flying straight and level, as on a patrol leg, the wing's task does not require the generation of many outputs. In Figure 11, these periods of little activity are evident in the periodic flat segments in the wing's cumulative outputs. When the lead begins a maneuver (e.g., a turn), the wing must maintain the formation throughout the maneuver. During a turn the wing generates many motor commands as it follows the lead. Because the turn takes a few seconds to complete, the outputs increase gradually over the course of the turn, as can be seen in the figure. Thus, the wing periodically encounters a dynamic situation that requires significant reasoning and motor responses. Further, the response to this change is not discrete, as in the lead, but occurs continuously over the course of the lead's maneuver.

These differences in the tasks for the two agents account for the relatively large absolute differences in the performance metrics between the lead and wing agents. Because the wings are adjusting their positions relative to the leads, they issue many more output commands than the leads, which requires many more inferences to determine what those commands should be.

**Decisions**: The differences between decisions in the lead and wing is due to an artifact of the data collection. The lead agents ran for an extra second after the wings halted in order to initiate data collection.

**Production Firings**: In both the lead and wing agents, production firings decrease. However, the wing's production firings decrease by 62%, while in the lead, the decrease is only 15%. One reason for the large improvement in the DHJ wing is due to the elimination of some redundant output commands in the FHJ agents. The FHJ wing sometimes issues the same motor command more than once. The reason for this duplication is that a specific motor command is computed locally, and is thus not available to other subtasks. In some cases, two subtasks may issue the same motor command. When the command is stored locally, a command may be issued again because the agent cannot recognize that the command has already been issued by another subtask. Because motor commands are remembered in the top subtask by DHJ agents, they can be inspected by all subtasks. The DHJ wing thus never issues a redundant motor command. The large relative decrease in outputs in the wing agent from FHJ to DHJ (Figure 11) can be attributed to this improvement. Production firings decrease with the decrease in output activity because most reasoning activity in the wing concerns reacting to the lead's maneuvers.

In contrast to the wing, the lead's average number of outputs actually increases. Regeneration is the source of these additional outputs. In a few situations, a DHJ agent's subtask for adjusting heading, speed or altitude can get updated repeatedly in a highly dynamic situation (e.g., a hard turn). The FHJ agent uses subtask knowledge to decide if the current output command needs to be updated. However, in DHJ, the subtask may be retracted due to dependence on a changing value (e.g., current heading). When the subtask is regenerated following a retraction, the lead may generate a slightly different motor command. For example, the lead might decide to turn to heading $90.1°$ instead of $90.2°$. This decision causes the generation of a new output command that would not have been re-issued in FHJ agents and accounts for the small increase in outputs. It also suggests that without the self-imposed constraint of the methodology, the knowledge base could be further modified to avoid this regeneration and further decrease production firings.

Although the large magnitude of the improvement in the wing is primarily due to remembering motor commands, both agents also needed less consistency knowledge and thus accessed less knowledge while performing the same task. The agents perform the same tasks using less knowledge.

**CPU Time**: CPU time decreases in both the DHJ lead and wing agents. The improvement in the lead (39%) is about half the improvement in the wing (81%). These differences are due primarily to the decrease in production firings. There are fewer production firings and thus fewer instantiations to generate, leading to improvements in CPU time. Match time also improved, contributing to the overall performance improvement.[9] The larger improvements in CPU time as compared to production firings improvements (39% vs. 15% in the lead, 81% vs. 62% in the wing) might be attributable to decreases in both the number of rule firings and match time. Again, these results offer no guarantee that match time will always decrease with DHJ. It is important to note, however, in two very different domains DHJ reduces total knowledge and further constrains the remaining knowledge. The architecture then leveraged these small differences for improved overall performance.

### 4.4.5 DIFFERENCES IN RESPONSIVENESS

Because CPU time decreases in the DHJ agents, responsiveness should generally improve. However, because some agent knowledge has been split into several different subtasks, some actions may not be initiated as quickly as would be initiated by the FHJ agent. In this section, we explore differences in responsiveness in one of these situations.

---

9. As in the Dynamic Blocks World, these trends are based on a few observations of the data, rather than a significant analysis. In particular, in $\mu$TAS, data for the number of tokens generated was not collected. The results reported here are consistent with the expectation that the token activity falls in DHJ agents, as compared to FHJ agents.

|       | Avg. In-Range Time (sec) | Avg. Launch Time (sec) | Reaction Time (sec) | $n$ |
|-------|--------------------------|------------------------|---------------------|-----|
| FHJ   | 161.816                  | 162.084                | .268                | 95  |
| DHJ   | 162.048                  | 162.993                | .945                | 99  |

Table 4: A comparison of average reaction times for launching a missile in $\mu$TAS.

When an enemy plane comes in range, the agent executes a series of actions, leading to the firing of a missile. *Reaction time* is the difference between the time at which an enemy agent comes in range and the time when the agent actually pushes the fire button to launch a missile. This reaction time is one measure of the agent's responsiveness. As Table 4 shows, the FHJ agent is able to launch the missile in just over a quarter of a second. However, the DHJ agent is about three-and-a-half times slower than the FHJ agent in launching the missile, taking almost a full second, on average.

Split subtasks, regeneration, and subtask selection all contribute to the increase in reaction time. Splitting a subtask with $n$ steps, which may have all been executed in a single decision previously, may now take $n$ decisions in the DHJ agent. Only a few actions are necessary for launching a missile so one would expect an increase of, at most, a few hundred milliseconds for this change. However, by dividing subtasks into separate steps, the sequential series of actions can be interrupted. In particular, a number of regenerations occur under the `launch-missile` subtask as the agent prepares to fire the missile in a highly dynamic situation. The agent sometimes chooses to undertake a similar action because the situation has changed enough that a slightly different action might be necessary, as described above. The result is that the DHJ agents are taking more accurate aim than the FHJ agents, as they are responding more quickly to the dynamics of the environment. This "aiming," however takes more time, although the increase in time is not tactically significant (i.e., enemy planes were not escaping that were previously hit by FHJ agents).

Some additional re-engineering of the knowledge would improve the reaction time (e.g., as described in Section 3.4.3). However, decreases in responsiveness will be difficult to avoid, in general. Dynamic Hierarchical Justification requires that subtasks with different dependencies be initiated and terminated separately, or risk unnecessary regeneration. However, by splitting complex tasks into separate subtasks, individual actions are delayed both because the subtasks are now separate procedures, and because the selection for a particular subtask in the series can be postponed when additional subtask choices are available.

## 4.5 Summary of Empirical Evaluations

Figure 12 summarizes results from the Dynamic Blocks World and $\mu$TAS. In both domains, DHJ agents require fewer total productions, suggesting a decrease in knowledge cost. Performance is roughly the same in the Dynamic Blocks World and for the lead agents in $\mu$TAS. The DHJ wing agents show a much greater improvement in overall performance, which is due both to DHJ and to changes in knowledge. These results suggest that Dynamic Hierarchical Justification can be expected to reduce engineering effort and not degrade performance in a variety of domains, simple and complex. However, response time in some situations may decrease.

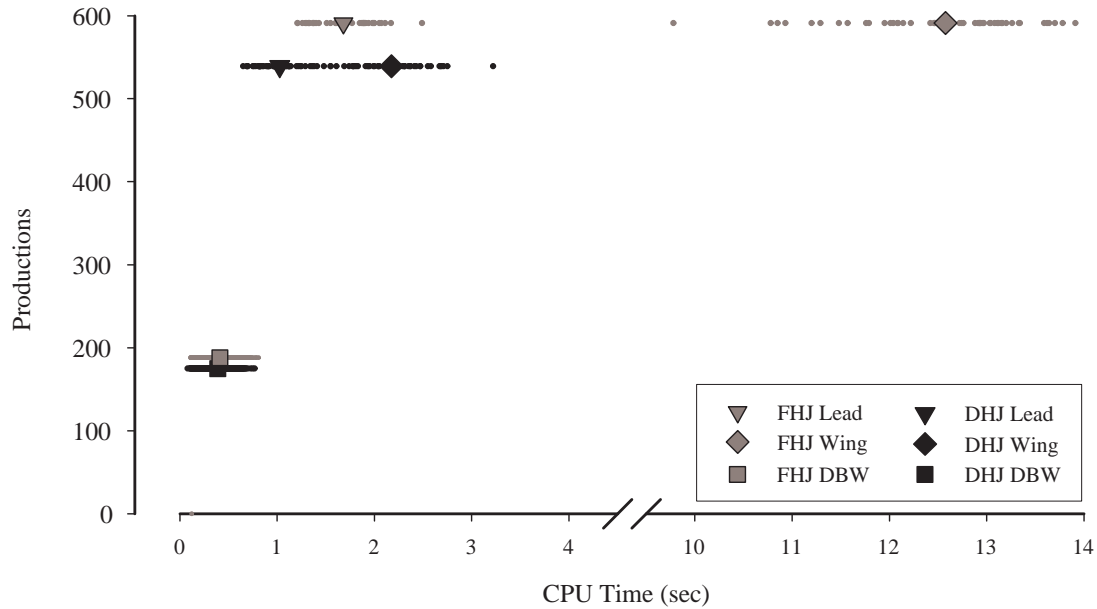

Figure 12: Mean CPU Time *vs.* knowledge in productions for FHJ (black) and DHJ (gray) agents in the Dynamic Blocks World and $\mu$TAS. The graph includes the actual distribution of CPU time for each agent as well as the mean for each agent. Means for the Dynamic Blocks World agents are illustrated with squares, $\mu$TAS lead agents with triangles, and $\mu$TAS wing agents with diamonds.

## 5. Discussion

Solutions other than KBAC and the new solutions introduced here can be developed for the inconsistency problem (Wray, 1998). We briefly introduce a few additional solutions, and also consider the relationship of Dynamic Hierarchical Justification to intention reconsideration in belief-desire-intention agents and belief revision.

### 5.1 Other solutions to inconsistency across the hierarchy

In this section, we review other existing architectural solutions to the problem of inconsistency arising from persistence in a hierarchy of assertions.

#### 5.1.1 Limiting Persistence

One obvious approach to eliminating inconsistency arising from persistence is to disallow persistent assumptions altogether. This approach was adopted in Theo (Mitchell et al., 1991). All reasoning in Theo is entailed from sensors; only perceptual inputs are unjustified. Theo cannot reason non-monotonically about any *particular* world state; only the world can change non-monotonically. Thus, Theo cannot generally remember previous inputs.

Another possible limitation would be to restrict all assumptions to a single memory (global state), or, equivalently, allow assumptions only in the root level of the hierarchy in a hierarchical architecture. This solution ensures that the hierarchical context is always consistent (all assertions within an associated subtask are entailments) and also allows persistence. Because HTN execution systems such as RETSINA and DECAF, mentioned previously, have only a global state, they obviously do not suffer from inconsistency over a hierarchy. However, the interactions between persistent assertions and new information derived from sensors will be a problem in systems with global state.

RETSINA has recently adopted *rationale-based monitoring* (Veloso, Pollack, & Cox, 1998) to identify environmental changes that could impact a currently executing task network (Paolucci et al., 1999). Rationale-based monitoring uses the structure of plan knowledge (in this case, plan operators, including task networks) to alleviate inconsistency. Monitors for relevant world features are created dynamically as planning progresses by identifying pre-conditions in operators and instantiating them via a straightforward taxonomy of monitor types (e.g., a monitor for quantified conditions). The collection of monitors form a *plan rationale*, "reasons that support a planner's decisions" (Veloso et al., 1998). Plan rationales are thus similar to the justifications used in truth maintenance. Monitors are activated when a pre-condition element in the world changes. They then inform the planner of the change, and the planner can then deliberate about whether the change should impact the plan under construction and, if so, consider appropriate repairs.

Rationale-based monitoring is similar to Dynamic Hierarchical Justification, especially because they both leverage the structures of their (different) underlying task representations to provide consistency. However, there are two important differences. First, because DHJ identifies the specific subtask impacted by a change, it does not require deliberation to determine the impact of the change; immediate return to a consistent knowledge state is possible. When a monitor is activated in rationale-based monitoring, the planner must first determine where and how the plan is affected, which can require deliberation. Second, because monitors trigger deliberation, rather than automatically retracting reasoning, an agent using rationale-based monitoring can determine if the plan should be repaired and how. DHJ (as implemented) does not offer this flexibility; retraction is automatic. Automatic retraction assumes the cost of retrieving (or regenerating) pre-existing plan knowledge is less costly than deliberation to determine if/how the plan can be revised. Because plan modification can be as expensive as plan generation (Nebel & Koehler, 1995), this assumption is reasonable. However, invoking a deliberate revision process could circumvent potential problems arising from recovery from interruption (Section 3.4.3).

### 5.1.2 Fixed Hierarchical Justification

As mentioned previously, both the pre-DHJ version of Soar and the Procedural Reasoning System (PRS) (Georgeff & Lansky, 1987) use Fixed Hierarchical Justification to retract complete levels of the hierarchy when the support set no longer holds. In PRS, the support set consists of a set of context elements that must hold during the execution of the subtask. These elements are defined by a knowledge engineer. Fixed Hierarchical Justification offers a complete solution to the inconsistency problem if context references within the reasoning in a subtask are limited to the support set. This approach guarantees consistency. However, it requires that the knowledge designer identify all potentially relevant features used in reasoning within the subtask. Additionally, the resulting system may be overly sensitive to the features in the support set if those features only rarely impact reasoning, leading to unnecessary regeneration.

Fixed Hierarchical Justification requires less explicit consistency knowledge than knowledge-based solutions. However, KBAC knowledge is still required if access to the whole task hierarchy is possible. Thus, an agent's ability to make subtask-specific reactions to unexpected changes in the environment is limited by the knowledge designer's ability to anticipate and explicitly encode the consequences of those changes.

## 5.2 Intention Reconsideration

In the belief-desire-intention (BDI) model of agency, an intention represents a commitment to achieving a goal (Rao & Georgeff, 1991; Wooldridge, 2000). An intention is thus similar to the instantiation of a subtask in a hierarchical architecture.

Dynamic Hierarchical Justification can be viewed as a partial implementation of *intention reconsideration* (Schut & Wooldridge, 2000, 2001). Intention reconsideration is the process of determining when an agent should abandon its intentions (due to goal achievement, recognition of failure, or recognition that the intention itself is no longer desired). Dynamic Hierarchical Justification is only a partial implementation of intention reconsideration because it is only able to capture syntactic features of the problem solving (i.e., the identification of dependencies via the support set) to determine when to reconsider an intention. Situations that require deliberation to determine that an intention should be abandoned are not captured in DHJ.[10] Schut & Wooldridge (2001) describe an initial attempt to allow the run-time determination of reconsideration policies. An optimal policy would maximize the likelihood that deliberate intention reconsideration actually leads to abandoning an intention (i.e., the agent reconsiders when reconsideration is necessary). In contrast, Dynamic Hierarchical Justification offers a low-cost, always available, domain general process for abandoning intentions, but cannot automatically identify reconsiderations requiring semantic analysis of the problem state.

In BDI models, agents can choose to execute their current action plans with or without reconsidering their current intentions first. Kinny and Georgeff (1991) showed that, in more static domains, "bold" agents that never reconsider their intentions perform more effectively than "cautious" agents that always reconsider before executing a plan step. The opposite is true in highly dynamic domains: "cautious" agents out perform "bold" ones. Both Soar and PRS can be described as being cautious via Fixed Hierarchical Justification. That is, at each plan step, the architectures determine if elements in the support set remain asserted before executing the step. FHJ approaches are, in effect, more "bold" than they appear, because they do not reconsider intentions when assertions have changed in the dependent context, but not in the support set. Dynamic Hierarchical Justification provides more "cautious" agents, because it ensures that the agent's reconsideration function takes into account all context dependencies for subtask reasoning. From the perspective of intention reconsideration, the problems introduced by more dynamic domains prompted us to explore more "cautious" solutions.

---

10. DHJ does not preclude deliberate reconsideration. However, Soar (as the testbed for the exploration of DHJ) does not provide an architectural solution for deliberate reconsideration. Thus, these situations will be addressed through knowledge and the deliberative processes of the architecture.

The results of the empirical analysis were somewhat consistent with those of Kinny & Georgeff. The "cautious" DHJ agents performed better than less "cautious" FHJ agents in the highly dynamic $\mu$TacAir-Soar domain. In the Dynamic Blocks World, the performance differences were more equivocal. In comparison to FHJ the number of new intentions increased with Dynamic Hierarchical Justification (measured as Soar decisions). While there was a slight overall performance improvement with DHJ, it was due to improvements in the match time of productions, a Soar-specific measure that likely will not generalize to other systems. These results suggest that DHJ is possibly overly cautious in static domains. However, because Dynamic Hierarchical Justification did not present a significant performance cost and unexpectedly played a constructive role in agent execution even in the static domain, DHJ seems warranted in both static and dynamic domains.

## 5.3 Belief Revision

Belief Revision refers to the process of changing beliefs to accommodate newly acquired information. The inconsistency problem is an example of the need for revision in asserted beliefs: some change in the hierarchical context (deriving ultimately from perceived changes in the world) leads to a situation in which a currently asserted assumption would not (necessarily) be regenerated if it were re-derived. Theories of belief revision identify functions that can be used to update a belief set so that it remains consistent.

The best known theory of belief revision is the "AGM" theory (Alchourrón, Gärdenfors, & Makinson, 1985; Gärdenfors, 1988, 1992). AGM is a *coherence* theory, meaning that changes to beliefs are determined based on mutual coherence with one another. This approach contrasts with the *foundations* approach, in which justifications (reasons) determine when/how to revise a belief set. Obviously, Dynamic Hierarchical Justification is an extension to the foundations approach to belief revision. However, as the foundations and coherence approaches can be reconciled (Doyle, 1994), in this section we explore the repercussions of Dynamic Hierarchical Justification in the context of the AGM theory of belief revision.

In AGM theory, when a new sentence is presented to a database of sentences representing the current knowledge state, an agent is faced with the task of revising its knowledge base via one of three processes: expansion (adding sentences to the knowledge base), contraction (removing sentences from the knowledge base) and revision (a combination of expansions and contractions). AGM theory emphasizes making minimal changes to a knowledge base and epistemic entrenchment, a notion of the usefulness of a sentence within the database. AGM theory prefers that sentences with high epistemic entrenchment (relative to other sentences) are retained during revision.

Comparing Dynamic Hierarchical Justification to Assumption Justification suggests that it is sometimes cheaper to remove a subtask (and all asserted beliefs associated with that subtask) than it is to compute the minimal revision with Assumption Justification. In the context of belief revision, this result is not surprising, since it has been shown that computing a minimal revision to a knowledge base can be computationally harder than deduction (Eiter & Gottlob, 1992). This theoretical result has led to applications that compute belief updates via incremental derivations of a belief state, rather than via belief revision (Kurien & Nayak, 2000).

The power of the heuristic approach used by DHJ over the analytic solution follows from the characteristics outlined in Section 3.4.1: the hierarchical structure and organization of the agent assertions and the efficiency of the underlying reasoning system to regenerate any unnecessarily removed assertions. Assumptions (persistent beliefs) are associated with particular subtasks in hierarchical architectures. A change in perception (an epistemic input) leads to a revision. Rather than determining the minimal revision, DHJ uses a heuristic that, in this context, says that persistent beliefs in a subtask have similar epistemic entrenchment to the subtask/intention itself. In some cases, this heuristic will be incorrect, leading to regeneration, but, when correct, it provides a much simpler mechanism for revision. Gärdenfors (1988) anticipates such conclusions, suggesting that systems possessing additional internal structure (as compared to the the relatively unstructured belief sets of AGM theory) may provide additional constraints for orderings of epistemic entrenchment.

## 6. Conclusion

The empirical results from both the Dynamic Blocks World and $\mu$TAS domains were consistent with expectations: knowledge engineering cost decreased and overall performance in DHJ was roughly the same (or slightly improved) in comparison to independently-developed FHJ benchmarks. Development cost decreases because the designer is freed from the task of creating across-level consistency knowledge. One drawback of DHJ is that responsiveness can degrade when regeneration occurs.

DHJ has been incorporated into the currently released version of Soar (Soar 8) for over 3 years and the experience of users further confirms that development cost decreases. It is partly true that developers need a deeper understanding of the architecture to realize this benefit. However, DHJ removes the need for the encoding of across-level consistency knowledge, which has proven difficult to understand and encode in many systems. DHJ also makes understanding the role of assumptions in Soar systems more straightforward, by imposing design and development constraints. For instance, the knowledge designer must now think about why, when, and where persistence should be used in the agent. Once the knowledge designer determines the functional role of some persistent assumption, DHJ guides the development of the knowledge necessary for that assumption. For a nonmonotonic or hypothetical assumption, no knowledge must be created that "looks outside" the subtask in order to ensure consistency (i.e., no across-level knowledge is necessary). Assumptions for remembering must be asserted in the root level of the hierarchy, and knowledge must be created to manage the remembered assumption. Functions of the root task now include monitoring, updating, and removing remembered assumptions (we are developing domain-general methods for managing these remembered assumptions to further reduce cost). Thus, while DHJ does increase the complexity of the architecture, it makes design decisions more explicit and manageable than previous KBAC approaches.

Regeneration, seemingly one of the drawbacks of DHJ, also contributes to decreased knowledge development costs. Regeneration serves as a debugging tool, allowing immediate localization of problem areas in the domain knowledge (and its specific decomposition). This debugging aid contrasts with previous knowledge development in which inconsistency often became evident only in irrational behavior, making it often difficult to determine the actual source of a problem. Thus, in addition to reducing total knowledge necessary for some task, Dynamic Hierarchical Justification might also reduce the cost per knowledge unit when creating agent knowledge by localizing problems via regeneration. However, if it is the case that some domain cannot be decomposed into nearly decomposable subunits, regeneration could be debilitating.

Another positive consequence of DHJ is that an agent may behave more robustly in novel situations not anticipated by the knowledge engineer. For example, as a simple experiment, the FHJ and DHJ Dynamic Blocks World agents were placed in the situation described in Figure 3. The FHJ agent fails when the block moves because it lacks knowledge to recognize moving blocks; the knowledge designer assumed a static domain. With the same knowledge, however, the DHJ agent responds to this situation gracefully. In the specific situation in Figure 3, the DHJ agent immediately retracts the `put-on-table(3)` subtask, because **block-3** is on the table, and thus the selection of that subtask is no longer consistent with the current situation. The agent then chooses `stack(2,3)` and decomposes this subtask into actions to put **block-2** on **block-3**. If a new block (e.g., **block-4**) is placed in the **empty** space below **block-2**, the architecture responds by retracting the subtask goal for `put-down(2)` (i.e., the subtask that contains the **empty** assumption). It then begins to search for empty spaces in order to continue its attempt to put **block-2** on the table. Because the architecture, rather than agent knowledge, ensures consistency across the hierarchy, DHJ agents should be less brittle in situations not explicitly anticipated by agent design.

DHJ also provides a solution to the problem of learning rules with non-contemporaneous constraints (Wray, Laird, & Jones, 1996). Non-contemporaneous constraints arise when temporally distinct assertions (e.g., red light, green light) are collected in a single learned rule via knowledge compilation. A rule with non-contemporaneous constraints will not lead to inappropriate behavior but rather will never apply. This problem makes it difficult to use straightforward explanation-based learning approaches to operationalize agent execution knowledge. Non-contemporaneous constraints arise when the architecture creates persistent assumptions that can become inconsistent with the hierarchical context (Wray et al., 1996). Because DHJ never allows such inconsistency, it solves the non-contemporaneous problem. For instance, agents in both the Dynamic Blocks World and $\mu$TAS were able to learn unproblematically in the new architecture, with no/little knowledge re-design. Wray (1998) provides additional details and an empirical assessment of the learning.

Dynamic Hierarchical Justification operates at a higher level of granularity than Assumption Justification or knowledge-based solution methods, trading fine-grained consistency for lower computational cost. This higher level of abstraction does introduce additional cost in execution. In particular, necessary regeneration led to some redundancy in knowledge search in both the Dynamic Blocks World and $\mu$TAS agents. Although overall efficiency improved, some of the improvement was due to improvements in the average match cost of productions, which cannot be guaranteed in all domains or in other architectures. Further, Dynamic Hierarchical Justification requires that complex subtasks be split into distinct subtasks. This requirement improves the knowledge decomposition and reduces regeneration in performance but can reduce responsiveness. However, with the straightforward compilation of reasoning in subtasks that DHJ enables, the reduction in responsiveness can be overcome with learning (Wray, 1998).

Although the implementation and evaluation of DHJ was limited to Soar, we attempted to reduce the specificity of the results to Soar in two ways. First, we identified the problems that across-level consistency knowledge introduces in knowledge-based approaches: it is expensive to develop, degrades the modularity and simplicity of the hierarchical representation, and is only as robust as the knowledge designer's imagination. When agents are developed in sufficiently complex domains, the expense of creating this knowledge will grow prohibitive. This cost may lead additional researchers to consider architectural assurances of consistency. Second, Dynamic Hierarchical Justification gains its power via the structure of hierarchically decomposed tasks. Although specific implementations may differ for other agent architectures, the heuristic simplifications employed by DHJ should transfer to any architecture utilizing a hierarchical organization of memory for task decomposition. Dynamic Hierarchical Justification is an efficient, architectural solution that ensures reasoning consistency across the hierarchy in agents employing hierarchical task decompositions. This solution allows agents to act more reliably in complex, dynamic environments while more fully realizing low cost agent development via hierarchical task decomposition.

## Acknowledgments

This work would not have been possible without those who contributed directly to the development and evaluation of Dynamic Hierarchical Justification. Scott Huffman, John Laird and Mark Portelli implemented Assumption Justification in Soar. Ron Chong implemented a precursor to DHJ. Randy Jones, John Laird, and Frank Koss developed $\mu$TacAir-Soar. Sayan Bhattacharyya, Randy Jones, Doug Pearson, Peter Wiemer-Hastings, and other members of the Soar group at the University of Michigan contributed to the development of the Dynamic Blocks World simulator. The anonymous reviewers provided valuable, constructive comments on earlier versions of the manuscript. This work was supported in part by a University of Michigan Rackham Graduate School Pre-doctoral fellowship, contract N00014-92-K-2015 from the Advanced Systems Technology Office of DARPA and NRL, and contract N6600I-95-C-6013 from the Advanced Systems Technology Office of DARPA and the Naval Command and Ocean Surveillance Center, RDT&E division. Portions of this work were presented at the $15^{th}$ National Conference on Artificial Intelligence in Madison, Wisconsin.

## Appendix A: Improving Task Decompositions

This appendix describes in detail the changes that were made to the $\mu$TAS agent knowledge for DHJ.

**Remembering:** Figure 4 showed an agent computing a new heading as a subtask of the `achieve-proximity` subtask. This calculation usually depends upon the current heading. When the agent generates the command to turn, the heading changes soon thereafter. In this situation, the DHJ agent must "remember" that it has already made a decision to turn to a new heading by placing the assumption that reflects the new heading in the top level. If it places the assumption in the local level, then the new current heading will trigger the removal of `turn-to-heading` and then regeneration of the subtask (if the agent determines that it still needs to turn to some new heading).

In the FHJ agents, all output commands (such as turn to some specific heading) were asserted as assumptions in the local subtask. The DHJ agent's knowledge was changed to issue output commands directly to the output interface (which, in Soar, is always part of the highest subtask in the hierarchy). No unnecessary regeneration now occurs because the agent remembers all motor commands and generates a new one only when a different output is necessary. This change, of course, requires consistency knowledge because the motor commands are unjustified and thus must be explicitly removed, as is true for any remembered knowledge with DHJ.

**Within-level Consistency Knowledge:** Dynamic Hierarchical Justification, like all solutions to the across-level consistency problem, still requires consistency knowledge within an individual subtask. Some of this knowledge in the FHJ agents is used to remove intermediate results in the execution of a subtask. This "clean up" knowledge allows the agent to remove local assertions that contributed to some terminating subtask and thus avoid the (mis)use of these assertions in later reasoning.

As an example, consider the `achieve-proximity` subtask. This subtask is used in a number of different situations when an agent needs to get closer to another agent. If the wing strays too far from the lead, it may invoke `achieve-proximity` to get back into formation with the lead. The lead uses `achieve-proximity` to get close enough to an enemy aircraft to launch a missile. The subtask requires many local computations as the agent reasons about what heading it should take to get closer to another aircraft. The specific computation depends on what information is available about the other aircraft. When the wing is pursuing the lead, it may know the lead's heading and thus calculate a collision course to maximize the rate of convergence. Sometimes the other agent's heading is not available. In this case, the agent simply moves toward the current location of the other agent. These local computations are stored in the local subtask. When `achieve-proximity` is terminated in the FHJ agent, the agent removes the local structure. Removing the structure is important both because it interrupts entailment of the local structure (e.g., calculation of the current collision course) and guarantees that if the agent decides to `achieve-proximity` with a different aircraft, supporting data structures are properly initialized. This knowledge thus maintains consistency in the local subtask by removing the local structure when the `achieve-proximity` subtask is no longer selected.

The FHJ agent could recognize when it was going to remove a subtask. The termination conditions in FHJ agents acted as a signal to the within-level consistency knowledge. The knowledge that removes the local structure for `achieve-proximity` can be summarized as: "if the `achieve-proximity` operator is selected, but its initiation conditions no longer hold, then remove the local `achieve-proximity` data structure." Thus, the FHJ agent uses a recognition of an inconsistency in the assertions to trigger the activation of this within-level consistency knowledge.

When the subtask's initiating conditions are no longer supported in the DHJ agents, the selected subtask is removed immediately. Thus, the DHJ agent never has the opportunity to apply the FHJ agent's within-level consistency knowledge. The failure to utilize this knowledge led to a number of problems, including more regenerations than expected.

To solve this problem, the local subtask data structure was created as an entailment of the initiation conditions of the subtask itself. When the subtask initiation conditions no longer held, both the subtask selection and the local structure are immediately removed by the architecture, requiring no additional knowledge. Thus, this change obviated the need for some within-level consistency knowledge. However, the local data structure may need to be regenerated if a subtask is temporarily displaced. For instance, the FHJ within-level consistency knowledge could determine under what conditions the local structure should be removed. The DHJ solution has lost that flexibility.

**Subtasks with Complex Actions:** FHJ agents can execute a number of actions in rapid succession, regardless of any inconsistency in the local assertions. A single subtask operator can be initiated in a situation representing the conditions under which to apply the first action in a sequence, and terminated when the last step in the sequence has applied. If some intermediate step invalidates the initiation conditions, the subtask still executes the actions.

Consider the process of launching a missile. An actual missile launch requires only the push of a button, assuming that previous steps such as selecting the target and an appropriate missile have been accomplished beforehand. After pushing the fire button, the pilot must fly straight and level for a few seconds while the missile rockets ignite and launch the missile into flight. Once the missile has cleared the aircraft, the agent "supports" the missile by keeping radar contact with the target. In FHJ agents, the `push-fire-button` subtask includes both the act of pushing the fire button and counting while the missile clears the aircraft. These tasks have different and mutually exclusive dependencies. The initiation condition for `push-fire-button` requires that no missile is already launched. However, the subsequent counting requires monitoring the newly launched missile.

DHJ agents using the FHJ knowledge base always remove the `push-fire-button` subtask as soon as the missile is perceived to be in the air, interrupting the complete procedure. Regeneration of the `push-fire-button` subtask occurs because the agent never waits for the missile to clear and thus never realizes that the missile just launched needs to be supported. The DHJ agent unsuccessfully fires all available missiles at the enemy plane.

Pushing the fire button and waiting for the missile to clear are independent tasks which happen to arise in serial order in the domain. We enforced this independence by creating a new subtask, `wait-for-missile-to-clear`, which depends only on having a newly launched missile in the air. The DHJ agent now pushes the fire button, selects `wait-for-missile-to-clear` to count a few seconds before taking any other action, and then supports the missile if it clears successfully.

This solution reduces regeneration and improves behavior quality but it does have a non-trivial cost. Whenever a subtask is split, the effects of subtask actions no longer occur in rapid succession within a decision. Instead, the effect of the first subtask occurs in one decision, the effect of the second subtask in the second decision, etc. Thus, this solution can compromise responsiveness.

## References

Agre, P. E., & Horswill, I. (1997). Lifeworld analysis. *Journal of Artificial Intelligence Research*, *6*, 111–145.

Alchourrón, C. E., Gärdenfors, P., & Makinson, D. (1985). On the logic of theory change: Partial meet contraction and revision functions. *Journal of Symbolic Logic*, *50*(2), 510–530.

Allen, J. F. (1991). Time and time again. *International Journal of Intelligent Systems*, *6*(4), 341–355.

Altmann, E. M., & Gray, W. D. (2002). Forgetting to remember: The functional relationship of decay and interference. *Psychological Science*, *13*, 27–33.

Bresina, J., Drummond, M., & Kedar, S. (1993). Reactive, integrated systems pose new problems for machine learning. In Minton, S. (Ed.), *Machine Learning Methods for Planning*, pp. 159–195. Morgan Kaufmann, San Francisco, CA.

Dechter, R. (1990). Enhancement schemes for constraint processing: Backjumping, learning and cutset decomposition. *Artificial Intelligence*, *41*, 273–312.

Doyle, J. (1979). A truth maintenance system. *Artificial Intelligence*, *12*, 231–272.

Doyle, J. (1994). Reason maintenance and belief revision. In Gärdenfors, P. (Ed.), *Belief Revision*, pp. 29–51. Cambridge University Press, Cambridge, UK.

Eiter, T., & Gottlob, G. (1992). On the complexity of propositional knowledge base revision, updates, and counterfactuals. *Artificial Intelligence*, *57*, 227–270.

Erol, K., Hendler, J., & Nau, D. S. (1994). HTN planning: Complexity and expressivity. In *Proceedings of the 12$^{th}$ National Conference on Artificial Intelligence*, pp. 1123–1128.

Firby, R. J. (1987). An investigation into reactive planning in complex domains. In *Proceedings of the 6$^{th}$ National Conference on Artificial Intelligence*, pp. 202–206.

Forbus, K. D., & deKleer, J. (1993). *Building Problem Solvers*. MIT Press, Cambridge, MA.

Forgy, C. L. (1979). *On the Efficient Implementation of Production Systems*. Ph.D. thesis, Computer Science Department, Carnegie-Mellon University.

Gärdenfors, P. (1988). *Knowledge in Flux: Modeling the Dynamics of Epistemic States*. MIT Press, Cambridge, MA.

Gärdenfors, P. (1992). Belief revision. In Pettorossi, A. (Ed.), *Meta-Programming in Logic*. Springer-Verlag, Berlin, Germany.

Gaschnig, J. (1979). Performance measurement and analysis of certain search algorithms. Tech. rep. CMU-CS-79-124, Computer Science Department, Carnegie-Mellon University, Pittsburgh, Pennsylvania.

Gat, E. (1991a). Integrating planning and reacting in a heterogeneous asynchronous architecture for mobile robots. *SIGART BULLETIN*, *2*, 71–74.

Gat, E. (1991b). *Reliable, Goal-directed Control of Autonomous Mobile Robots*. Ph.D. thesis, Virginia Polytechnic Institute and State University, Blacksburg, VA.

Georgeff, M., & Lansky, A. L. (1987). Reactive reasoning and planning. In *Proceedings of the 6$^{th}$ National Conference on Artificial Intelligence*, pp. 677–682.

Graham, J., & Decker, K. (2000). Towards a distributed, environment-centered agent framework. In Wooldridge, M., & Lesperance, Y. (Eds.), *Lecture Notes in Artificial Intelligence: Agent Theories, Architectures, and Languages VI (ATAL-99)*. Springer-Verlag, Berlin.

Hanks, S., Pollack, M., & Cohen, P. R. (1993). Benchmarks, test beds, controlled experimentation and the design of agent architectures. *AI Magazine, 14*, 17–42.

Hayes-Roth, B. (1990). An architecture for adaptive intelligent systems. In *Workshop on Innovative Approaches to Planning, Scheduling and Control*, pp. 422–432.

Jones, R. M., Laird, J. E., Neilsen, P. E., Coulter, K. J., Kenny, P., & Koss, F. V. (1999). Automated intelligent pilots for combat flight simulation. *AI Magazine, 20*(1), 27–41.

Kinny, D., & Georgeff, M. (1991). Commitment and effectiveness of situated agents. In *Proceedings of the 12th International Joint Conference on Artificial Intelligence*, pp. 82–88.

Kurien, J., & Nayak, P. P. (2000). Back to the future for consistency-based trajectory tracking. In *Proceedings of the 17th National Conference on Artificial Intelligence*, pp. 370–377.

Laird, J. E. (2001). It knows what you are going to do: Adding anticipation to a Quakebot. In *Proceedings of the 5th International Conference on Autonomous Agents*, pp. 385–392.

Laird, J. E., Congdon, C. B., & Coulter, K. J. (1999). Soar user's manual version 8.2. Manual, Department of Electrical Engineering and Computer Science, University of Michigan, http://ai.eecs.umiuch.edu/soar/docs.html.

Laird, J. E., Newell, A., & Rosenbloom, P. S. (1987). Soar: An architecture for general intelligence. *Artificial Intelligence, 33*, 1–64.

Laird, J. E., & Rosenbloom, P. S. (1990). Integrating execution, planning, and learning in Soar for external environments. In *Proceedings of the 8th National Conference on Artificial Intelligence*, pp. 1022–1029.

Laird, J. E., & Rosenbloom, P. S. (1995). The evolution of the Soar cognitive architecture. In Steier, D., & Mitchell, T. (Eds.), *Mind Matters: Contributions to Cognitive and Computer Science in Honor of Allen Newell*. Lawrence Erlbaum Associates, Hillsdale, NJ.

McDermott, D. (1991). A general framework for reason maintenance. *Artificial Intelligence, 50*, 289–329.

Mitchell, T. M., Allen, J., Chalasani, P., Cheng, J., Etzioni, O., Ringuette, M., & Schlimmer, J. C. (1991). Theo: A framework for self-improving systems. In VanLehn, K. (Ed.), *Architectures for Intelligence*, chap. 12, pp. 323–355. Lawrence Erlbaum Associates, Hillsdale, NJ.

Mitchell, T. M. (1990). Becoming increasingly reactive. In *Proceedings of the 8th National Conference on Artificial Intelligence*, pp. 1051–1058.

Nebel, B., & Koehler, J. (1995). Plan reuse versus plan generation: A theoretical and empirical analysis. *Artificial Intelligence, 76*, 427–454.

Newell, A. (1990). *Unified Theories of Cognition*. Harvard University Press, Cambridge, MA.

Paolucci, M., Shehory, O., Sycara, K. P., Kalp, D., & Pannu, A. (1999). A planning component for RETSINA agents. In Wooldridge, M., & Lesperance, Y. (Eds.), *Lecture Notes in Artificial Intelligence: Agent Theories, Architectures, and Languages VI (ATAL-99)*, pp. 147–161, Berlin. Springer-Verlag.

Pearson, D. J., Huffman, S. B., Willis, M. B., Laird, J. E., & Jones, R. M. (1993). A symbolic solution to intelligent real-time control. *Robotics and Autonomous Systems*, *11*, 279–291.

Rao, A. S., & Georgeff, M. P. (1991). Modeling rational agents within a BDI-architecture. In *Proceedings of the 2nd International Conference on Principles of Knowledge Representation and Reasoning*, pp. 471–484.

Russell, S., & Norvig, P. (1995). *Artificial Intelligence: A Modern Approach*. Prentice Hall, Upper Saddle River, NJ.

Sacerdoti, E. D. (1975). The nonlinear nature of plans. In *Proceedings of the 4th International Joint Conference on Artificial Intelligence*, pp. 206–214.

Schut, M., & Wooldridge, M. (2000). Intention reconsideration in complex environments. In *Proceedings of the 4th International Conference on Autonomous Agents*, pp. 209–216.

Schut, M., & Wooldridge, M. (2001). Principles of intention reconsideration. In *Proceedings of the 5th International Conference on Autonomous Agents*, pp. 340–347.

Shoham, Y. (1993). Agent-oriented programming. *Artificial Intelligence*, *60*(1), 51–92.

Simon, H. A. (1969). *The Sciences of the Artificial*. MIT Press, Cambridge, MA.

Stallman, R. M., & Sussman, G. J. (1977). Forward reasoning and dependency-directed backtracking in a system for computer aided circuit analysis. *Artificial Intelligence*, *9*(2), 135–196.

Sycara, K., Decker, K., Pannu, A., Williamson, M., & Zeng, D. (1996). Distributed intelligent agents. *IEEE Expert*, *11*(6), 36–46.

Tambe, M. (1991). *Eliminating Combinatorics from Production Match*. Ph.D. thesis, Carnegie-Mellon University. (Also published as Technical Report CMU-CS-91-150, Computer Science Department, Carnegie Mellon University.).

Tambe, M., Johnson, W. L., Jones, R. M., Koss, F., Laird, J. E., Rosenbloom, P. S., & Schwamb, K. (1995). Intelligent agents for interactive simulation environments. *AI Magazine*, *16*(1), 15–39.

Veloso, M. M., Pollack, M. E., & Cox, M. T. (1998). Rationale-based monitoring for planning in dynamic environments. In *Proceedings of the 4th International Conference on Artificial Intelligence Planning Systems*, pp. 171–180.

Wilkins, D. E., Myers, K. L., Lowrance, J. D., & Wesley, L. P. (1995). Planning and reacting in uncertain and dynamic environments. *Journal of Experimental and Theoretical Artificial Intelligence*, *7*(1), 197–227.

Wooldridge, M. (2000). *Reasoning about Rational Agents*. MIT Press, Cambridge, MA.

Wray, R. E. (1998). *Ensuring Reasoning Consistency in Hierarchical Architectures*. Ph.D. thesis, University of Michigan. Also published as University of Michigan Technical Report CSE-TR-379-98.

Wray, R. E., & Laird, J. (1998). Maintaining consistency in hierarchical reasoning. In *Proceedings of the 15$^{th}$ National Conference on Artificial Intelligence*, pp. 928–935.

Wray, R. E., Laird, J., & Jones, R. M. (1996). Compilation of non-contemporaneous constraints. In *Proceedings of the 13$^{th}$ National Conference on Artificial Intelligence*, pp. 771–778.

Wray, R. E., Laird, J. E., Nuxoll, A., & Jones, R. M. (2002). Intelligent opponents for virtual reality trainers. In *Proceedings of the Interservice/Industry Training, Simulation and Education Conference (I/ITSEC) 2002*.